\documentclass[journal]{IEEEtran}

\ifCLASSINFOpdf
\else
\fi
\usepackage[
 font=footnotesize
 ]{subfig}
 
\usepackage{dsfont}

 \usepackage[pdftex]{graphicx}
   \graphicspath{{../png/}}
   \DeclareGraphicsExtensions{.png}

\usepackage{mwe}
\usepackage[nonumberlist,acronym]{glossaries}
\usepackage{nomencl}
\usepackage{comment}
\usepackage{tikz}

\usepackage{amsmath}
\usepackage{graphicx}
\usepackage{ctable} 
\usepackage{amsfonts}

\usepackage{siunitx}
\usepackage{hyperref}
\usepackage{pgfplots}
\usepackage[compact]{titlesec}
\titlespacing{\section}{2pt}{*2}{*2}
\titlespacing{\subsection}{1pt}{*1}{*1}
\titlespacing{\subsubsection}{0.5pt}{*0.5}{*0.5}
\usepackage[font=footnotesize,skip=7.5pt,belowskip=-10pt]{caption}

\usepackage{algpseudocode, algorithm}

\usepackage{array}
\usepackage{float}

\def\BibTeX{{\rm B\kern-.05em{\sc i\kern-.025em b}\kern-.08em
    T\kern-.1667em\lower.7ex\hbox{E}\kern-.125em}}

\DeclareSIUnit{\microsecond}{\micro s}
\DeclareSIUnit{\millisecond}{\milli s}
\definecolor{blue(pigment)}{RGB}{0.2, 0.2, 0.6}

 \definecolor{changes}{rgb}{0, 0, 0}

\makeglossaries
\glsaddallunused

\begin{document}

%
\title{RALACs: Action Recognition in Autonomous Vehicles using Interaction Encoding \\ and Optical Flow}






%
%
%

\author{Eddy Zhou, Alex Zhuang, Alikasim  Budhwani,  Owen Leather,
Rowan Dempster, Quanquan Li, Mohammad Al-Sharman,~\IEEEmembership{Member,~IEEE,}
        Derek~Rayside,~\IEEEmembership{Member,~IEEE,}
        and~William Melek,~\IEEEmembership{Senior~Member,~IEEE}

\thanks{E. Zhou, A. Budhwani, O. Leather, Mohammad Al-Sharman, and W. Melek are with the Mechanical and Mechatronics Engineering Department and with WATonomous (Waterloo autonomous vehicle team in the SAE AutoDrive Challenge. (Watonomous.ca)), University of Waterloo, ON, CA, e-mail:eddy.zhou@uwaterloo.ca, a2budhwa@uwaterloo.ca, oleather@uwaterloo.ca, mkalsharman@uwaterloo.ca; william.melek@uwaterloo.ca. $^*$Corresponding author: { mkalsharman@uwaterloo.ca}}

\thanks{A. Zhuang is with Cheriton School of Computer Science and with WATonomous, University of Waterloo, ON, CA, e-mail: a5zhuang@uwaterloo.ca}

\thanks{ Q. Li, R. Dempster, and D. Rayside are with the Department of Electrical and Computer Engineering and with WATonomous, University of Waterloo. University of Waterloo, ON, CA, e-mail:  quanquan.li@uwaterloo.ca; r2dempster@uwaterloo.ca, drayside@uwaterloo.ca.}}

%
%

\markboth{IEEE Journal of \LaTeX\ Class Files,~Vol.~14, No.~8, August~2023}%
{Shell \MakeLowercase{\textit{et al.}}: Bare Demo of IEEEtran.cls for IEEE Journals}
%



\maketitle

\begin{abstract}

When applied to autonomous vehicle (AV) settings, action recognition can enhance an environment model's situational awareness. This is especially prevalent in scenarios where traditional geometric descriptions and heuristics in AVs are insufficient. However, action recognition has traditionally been studied for humans, and its limited adaptability to noisy, un-clipped, un-pampered, raw RGB data has limited its application in other fields. To push for the advancement and adoption of action recognition into AVs, this work proposes a novel two-stage action recognition system, termed RALACs. RALACs formulates the problem of action recognition for road scenes, and bridges the gap between it and the established field of human action recognition. This work shows how attention layers can be useful for encoding the relations across agents, and stresses how such a scheme can be class-agnostic. Furthermore, to address the dynamic nature of agents on the road, RALACs constructs a novel approach to adapting Region of Interest (ROI) Alignment to agent tracks for downstream action classification. Finally, our scheme also considers the problem of active agent detection, and utilizes a novel application of fusing optical flow maps to discern relevant agents in a road scene. We show that our proposed scheme can outperform the baseline on the ICCV2021 Road Challenge dataset \cite{singh2022road} and by deploying it on a real vehicle platform, we provide preliminary insight to the usefulness of action recognition in decision making. \textcolor{changes}{The code is publicly available at \url{https://github.com/WATonomous/action-classification}}.

\end{abstract}




%

\IEEEpeerreviewmaketitle

\section{Introduction}
\label{section:intro}

\IEEEPARstart{I}{n} the push to build robust and explainable autonomous driving systems, all forms of situational awareness are key to improving vehicle safety and redundancy for widespread adoption \cite{huang2020learning, dempster2023real, petrillo2020secure, al2020sensorless, xia2022onboard}. Existing iterations of AV perception systems make heavy use of object detection and tracking \cite{feng2020deep, hu2022sim}. Despite its widespread use, only utilizing an object's location and kinetics impedes an ego vehicle's ability to make proper planning decisions \cite{singh2022road}. In many road situations, knowledge of an object's past trajectory alone may not be sufficient enough to correctly extrapolate the object's future behaviour \cite{kamale2023cautious}. 
\begin{figure}[h!]
    \centering
  \subfloat[\centering \textbf{Detection:} Active agents are recognized from raw RGB video]{%
       \includegraphics[width=0.49\linewidth]{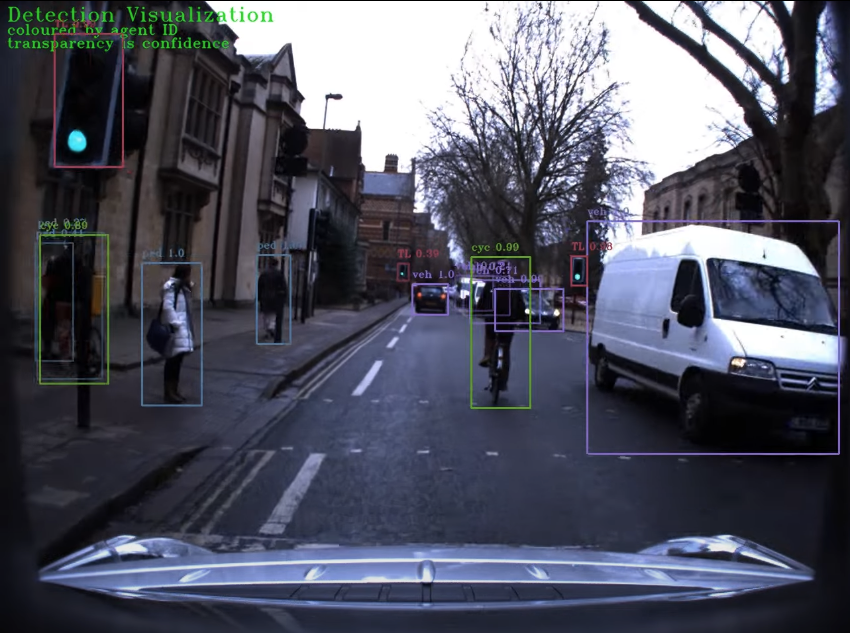}}
    \hfill
  \subfloat[\centering \textbf{Tracking:} Agent detections are temporally stitched together]{%
        \includegraphics[width=0.49\linewidth]{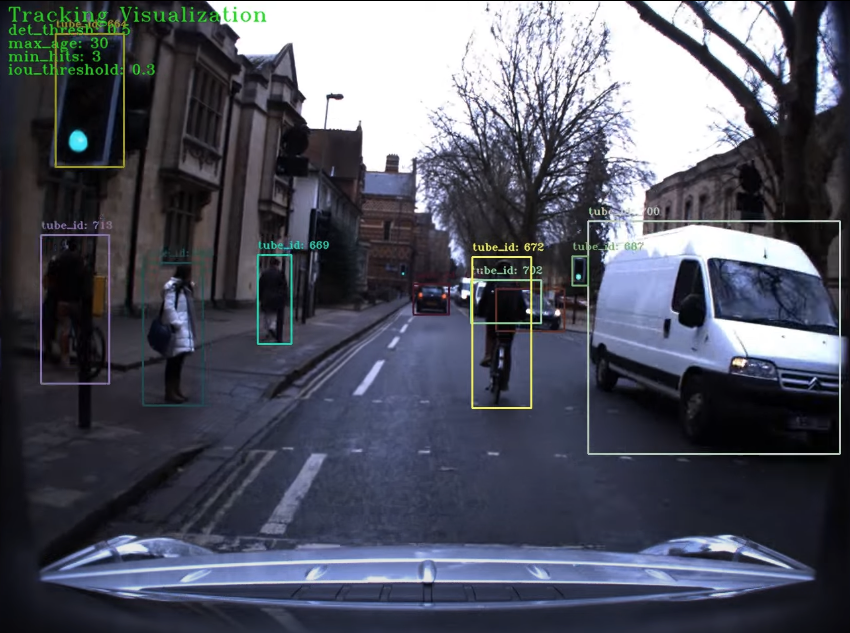}}
    \hfill
\\
  \subfloat[\centering \textbf{Action Attribute:} Agent tracks are appended with action confidence scores (highest score is colour of bounding box)]{%
        \includegraphics[width=0.7\linewidth]{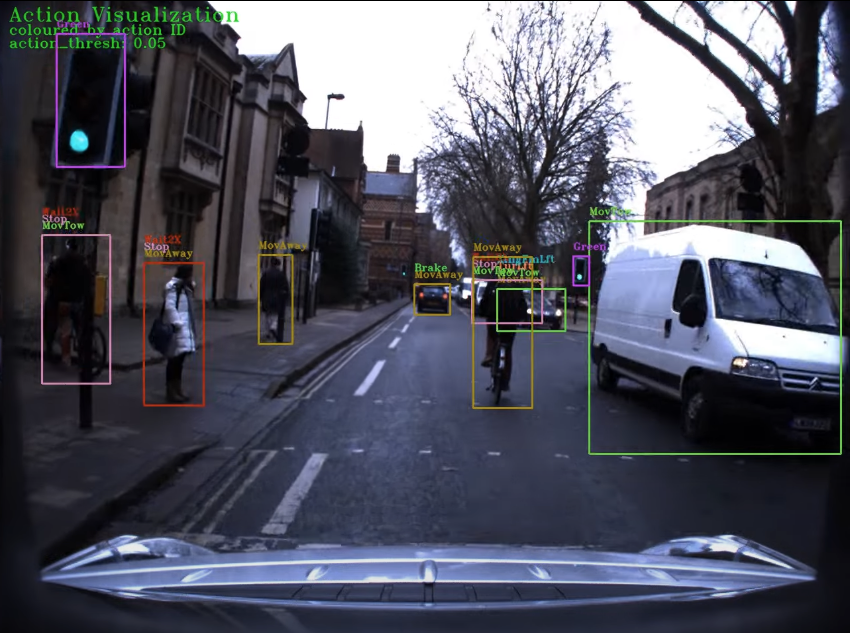}}
  \caption{Action Classification pipeline from raw RGB data. Agent tracks are appended with an action attribute which may be useful for downstream decision making. In (c) for example, we observe a pedestrian waiting to cross the street. Because of this attribute, our ego vehicle can have greater confidence in crossing the intersection, knowing that the pedestrian is properly obeying traffic rules. In this work, we focus our attention on making Action Classification more effective and realizable on autonomous vehicles, and we show how the use of encoding interactions between agents, and optical flow fusion detection, can significantly do this.}
  \label{fig:front} 
\end{figure} 
A pedestrian walking towards an intersection, for instance, may be predicted to cross regardless of what state the traffic indicators are at the time. As a result, having a separate set of action labels appended onto an agent, like \emph{waiting\_to\_cross} in Figure \ref{fig:front}, could be useful for downstream autonomous decision making.

Action localization and classification on the road pose a multitude of challenges. For one, sensor data coming from autonomous vehicles is not as perfectly clipped and pampered as those of human action datasets, and objects often enter and leave the scene within a matter of milliseconds. Furthermore, to make our system deployable in real time, it must have sub-second processing time along with little to no unnecessary processing. This means limiting the chances of running our system on objects that \textcolor{changes}{do not} have an action in the first place. 

It is entirely possible to use hardcoded rules to directly extract semantic queues to get an object's current action. Certain queues, like a car signalling left, could be found by traditional computer vision tools, but these queues can lead to multiple possible actions depending on context. As a result, extracting interactions between objects could be useful for more accurate action classification.

However, capturing all possible interactions between all road-relevant classes for action recognition through global heuristics is a daunting task. Hence, in this paper, we attempt to capture these interactions in the road domain using a deep neural framework. In other words, we aim to neurally encode the interactions between all possible combinations of road-relevant classes to produce action labels potentially useful for downstream autonomous decision making. 

Aside from classifying actions, localizing them pose its own challenge \cite{aboah2023deepsegmenter,li2023action}, especially when this paper centers around the use of action recognition systems in autonomous vehicles \textcolor{changes}{for the aforementioned reasons}. Consequently, being able to detect and track active agents in an online and robust manner for action recognition in autonomous vehicles is crucial. We found that optical flow maps generated by off-the-shelf neural architectures could be the key in differentiating active agents from stationary ones.

This paper lies in a unique position. On one hand, conventional action recognition methods demonstrate significant performance in human action datasets, but they require certain assumptions to be fulfilled. These include trimmed videos, object-centered clips, and idealized agent tracks. Furthermore, many top-performing action networks rely heavily on offline post-processing. On the other hand, action recognition for autonomous vehicles cannot meet these assumptions. Vehicles must make decisions in real time \cite{al2023self}, so action networks must not rely on ideal clips and offline post-processing.

\begin{figure*}[t]
  \centering
   \includegraphics[width=1\linewidth]{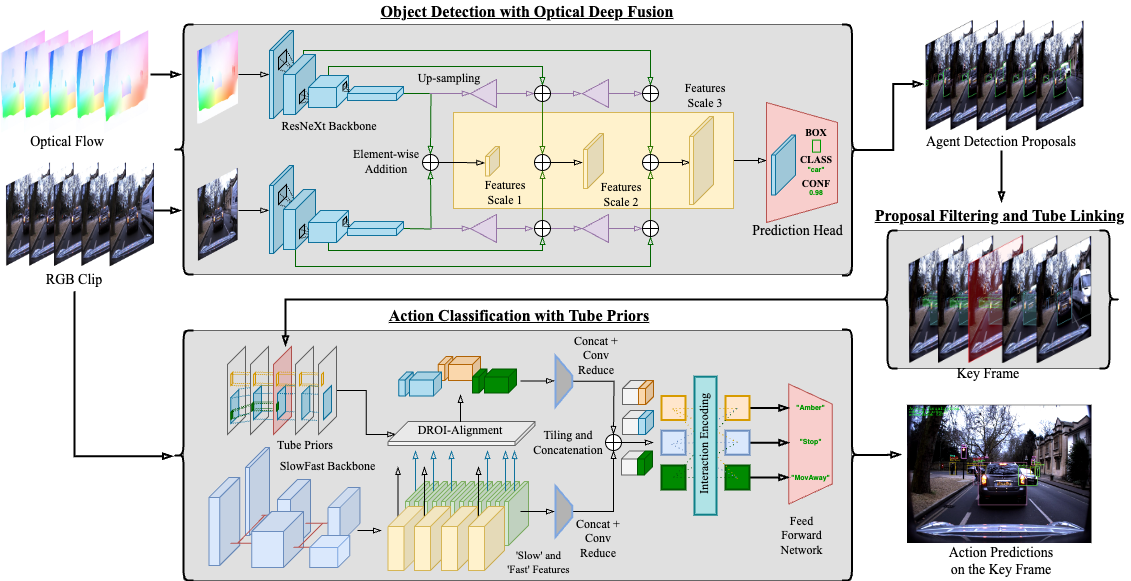}

   \caption{RALACs system architecture. Given a set of frames in a clip, their optical flow is estimated using RAFT \cite{teed2020raft}. Both clips are sent frame-by-frame into the object detector where each RGB frame and its corresponding optical flow is encoded and summed up at multiple feature scales. We found that utilizing a pretrained RGB backbone on 3-channel optical flow improved model predictions and accuracy. Following detection, an online object tracker is used to link detections into tubelets. Tubelets present in the key frame of the clip are then fed into our action classifier, which encodes agent tubes through a novel Dynamic ROI-Alignment procedure that takes advantage of the inherent structure of feature encodings outputted by SlowFast. Following encoding, we adopt \cite{acar2020}'s higher-order relations reasoning to compute the attention between each encoded agent tubelet and the other agents present in the clip. The result is the action predictions of agents present in the key frame.}
   \label{fig:fig4}
\end{figure*}

To address these problems, and sway research directions to bridge the gap between human action recognition and action recognition for the road, we propose the Road Active Agent Localization and Action Classification system (RALACs) (see Figure \ref{fig:fig4}), a novel scheme for road action recognition, following constraints that are closely tied to real autonomous vehicle implementation. Our main contributions include the following:
\begin{itemize}
    \item We adapt the idea of learning interactions between agents from previous works in human action classification and extend its capabilities to capture interactions between multiple road-relevant classes; not just humans and the environment.
    \item We formulate the problem of detecting active agents in a road scene and propose a novel action localization architecture that uses information from optical flow maps to detect only these agents.
    \item We formulate the problem of dynamic agents in high-motion scenes and \textcolor{changes}{propose a novel algorithm for Dynamic Region of Interest (DROI) alignment to encode agent paths in feature tensors which yields better results than existing single key-frame ROI alignment}.

\end{itemize}

To test RALACs' viability in autonomous vehicles, we share our initial attempt at implementing such a system on our autonomous vehicle research platform \cite{dempster2022drg}. We also provide preliminary insights to the possible advancements action labels can have on decision making. The remainder of the paper is structured  as follows. Section \ref{section:related-works} presents other related works. Section \ref{section:Methodology} illustrates the proposed action recognition methodology. Section \ref{section:experimentation} illustrates the experimental setup design, implementation details and the results of the proposed architecture followed by a discussion. Finally, section \ref{section:conclusion} outlines the proposed and future works.

\section{Related Works}

\label{section:related-works}

\subsection{Human Action Recognition}
 
Human action recognition is a rather mature field. The task of human action recognition can be split into two subtasks: action localization and action classification \cite{sun2021real,gao2021novel,wang2020recapnet}. Some models focus on processing each task separately \cite{li2013rectube,zhao2022tuber,acar2020} while others chose to both localize and classify actions at once by either transferring features from a localization step to a classifier \cite{song2019tac} or by doing both at once \cite{yowo2019,step2019}. Interestingly, numerous methods ignore the localization task as a whole \cite{li2021mvit,wu2022memvit,slowfast2018}. This is most likely due to the structure of action classification datasets, with some focusing more on global activity recognition in a clip \cite{hiel2015activity}. These action recognition models are incompatible for autonomous vehicles as they are not robust against untrimmed, unpampered video streams.

 As a safety-critical system, an autonomous vehicle should have explainable software modules that can assure the confidence of all stakeholders during widespread adoption.
 For the sake of developing deep neural models that are compatible and explainable with modular software in safety-critical systems, we focus our attention on designing a step-by-step approach to action localization and classification. Our related works reflect this.

\subsection{Human Action Localization}
 Action localization, or action proposal generation, consists of finding the temporal boundaries and spatial bounding boxes in which an active object exists.

 Most attempts at action localization focused purely on temporal localization using convolutional neural networks \cite{alwassel2021tsp,shou2016temporal} and LSTMs \cite{yeung2015every, escorcia2016dap}. These methods often followed an expensive sliding window approach \cite{laptev2007mov,tian2013spatial,oneata2014fish}.

 Existing spatial-temporal localization works consist of supervised, semi-supervised, and weakly supervised approaches. Many of these solutions use a sliding-window technique, with some increasing their accuracy offline. Recently, some works focus on building action tubelets incrementally, allowing for better use in real time systems. Our proposed action localization scheme follows a similar strategy.

\subsection{Optical Flow Estimation}
 Estimating optical flow is a difficult problem which was traditionally posed as the optimization of a global energy function \cite{Baker2011}. This optimization problem was solved using continuous \cite{Horn1981} and discrete optimization algorithms  \cite{Boykov2001}. Empowered by CNNs, estimating optical flows has become faster and robust using a simple two stream network, such as the FlowNetv2 architecture \cite{Ilg2016}. The FlowNetv2 design was improved upon by the RAFT \cite{teed2020raft} architecture which introduced the use of a recurrent operator that iteratively updates the flow field.

 The estimated optical flow can be used with raw RGB frames for various detection and segmentation tasks. \cite{Rashed2019}, \cite{Siam2017}, \cite{Mohamed2020}. \cite{Rashed2019} studied the use of optical flow with RGB frames for the task of semantic segmentation. The authors compared four different architectures and found that the two-stream architecture with deep fusion of RGB and optical flow resulted in better overall performance. We take inspiration from their approaches and adapt them for active agent detection.

\subsection{Action Recognition for Autonomous Vehicles}
 In contrast to human action, action recognition for autonomous vehicles is an under-explored field. The ROad event Awareness Dataset (ROAD) \cite{singh2022road} was the first to apply the concept of action classification to road agents, and a limited number of methods have been applied to this domain so far. In their description of the winning entries of different subtasks of the challenge, the ROAD authors emphasize that most performance improvements result from improvements in post-processing steps \cite{singh2022road}. This seems counter-intuitive for a dataset meant to cater action recognition for autonomous vehicles, which requires an accurate, online solution to action localization and classification. 
 
 In particular, the winning entry Argus++ \cite{yu2022argus++} is noted to be rather complex, proposing a tube proposal generation scheme that oversamples fixed length action tubes and subsequently deduplicates them. In contrast to their post-processing-focused approach, we choose to constrain ourselves to online approaches only.

\section{Methodology}
\label{section:Methodology}

In the following section, we provide a detailed explanation of the proposed RALACs system architecture illustrated in Figure \ref{fig:fig4}.

\label{section:methodology}

\subsection{Action Localization with Optical Flow}

Action localization consists of finding the spatial-temporal boundaries in which an active agent occurs. Compared to traditional object recognition, action localization systems must differentiate between agents with and without an action. This reduces unnecessary processing and the chances of misclassification of static road agents (eg. parked cars). 

Our approach focuses on the hypothesis that the relevance, or activeness, of an agent is related to its motion relative to the environment. We also follow an incremental, frame-based methodology to make our system implementable in real time.

For an arbitrary memory bandwidth of $l$ RGB frames back from current time, our action localization system produces a list of $n$ active agent tubelets $T = [T_1, ..., T_n]$. Each of these tubelets, $T_i$, contains the class of its active agent $c_i$ and that agent's bounding-boxes for each of the $l$ frames $[b_{i, 0}, ..., b_{i, l}]$. Since an active agent is subject to occlusion, and can disappear at any time, an active agent tublet may have no bounding-box for some of the frames.



\subsubsection{Optical-Flow-based Active Agent Detection}
In this section, we introduce a frame-wise approach to active agent detection using optical flow. To include motion information, we propose a modified, two-staged object detector which considers both the optical flow and RGB to make predictions. 

\textbf{Optical Flow Estimation:}
We generate optical flow maps using an online optical flow estimator such as RAFT\cite{teed2020raft}. RAFT contains several good attributes that make their network viable for autonomous vehicle applications. These include:

\begin{enumerate}
\item{
    Strong generalization ability based on the author's tests with synthetic data \cite{teed2020raft}.
}
\item{
    Fast inference speed of around $\sim$20 fps on an RTX 3090.
}
\end{enumerate}
\textcolor{changes}{Because of these attributes, we chose to use RAFT for generating our optical flow maps.}

\textcolor{changes}{A} persistent challenge with optical flow estimation for autonomous vehicles is that camera data is not coming from a stationary source. Consequently, we end up with noisy flow fields containing not just the motion of active agents but also stationary objects passing by as well. 

This issue can be addressed in two ways: 
\begin{enumerate}
\item{
    Filtering of flow vectors representing the flow magnitude of stationary objects. This would require hard-coded rules.
}
\item{
    Leaving the flow field as is.
}
\end{enumerate}

Although filtering stationary flow may seem promising, the process in which to filter it out changes according to the motion of the ego vehicle. For example, when the ego vehicle is moving straight ahead, flow vectors of stationary objects diverge from the center of the frame. In contrast, when the ego vehicle is turning, the stationary flow field may have little to no divergence at all. Instead, most of the stationary flow vectors are facing a direction opposite to which the ego vehicle is turning. There are many other cases that would cause the filtering process to change, so instead of relying on global heuristics, we intend to let our detection neural network implicitly learn to discount stationary flow using the inputted flow map and the RGB frame. 

\textbf{Fusion of Optical Flow with RGB:}
Our two-stream object detector must encode two data sources: the current RGB frame of size $h*w*3$ and the optical flow map, of size $h*w*2$, between the current frame and its previous. Because the optical flow map is different in size, we instead use a three-channel, color-wheel representation of optical flow. This can be done by normalizing the magnitudes of the flow vectors and mapping them to an image of a colour wheel. 

With these two sources, we propose a deep fusion, modified feature pyramid network (FPN) approach to encoding optical flow and RGB for frame-level, active-agent detection. 


We start by forwarding the optical flow map and RGB images separately into two different ResNeXt backbone networks. This produces multiple feature maps for each  data stream, denoted as $F_O = \{F_{O_1}, F_{O_2}, F_{O_3}, F_{O_4}, F_{O_5}\}$ for the feature maps of optical flow and $F_R = \{F_{R_1}, F_{R_2}, F_{R_3}, F_{R_4}, F_{R_5}\}$ for the feature maps of RGB. Note that the dimensions of $F_{O_i}$ and $F_{R_i}$, where $i$ is the feature level, are the same. Following feature mapping, $F_{O_i}$ and $F_{R_i}$ are then fused using a summation junction in the feature pyramid network (FPN) to form equation ( \ref{eqn:1}) or the final FPN feature scale for the $i$th level. Each $F_i$ is then fed into a Faster R-CNN prediction head, producing $m$ post-nms detections for a single frame $D=[(c_1, b_1), ..., (c_m, b_m)]$ where $c_i$ is the class of the agent with its confidence score and $b_i$ is its bounding box. We use a low nms threshold on the list of detections to increase the number of candidates for downstream tracking.  
\begin{equation}
\label{eqn:1}
    F_i = F_{O_i} + F_{R_i} + UpSample(F_{O_{i-1}}) + UpSample(F_{R_{i-1}})
\end{equation}
During training, the use of a summation junction makes backpropagation simple and intuitive. Since we use the three-channel RGB color-wheel representation of optical flow, we can use the same initial pretrained weights for both backbones. This was found to be better than starting the optical-flow backbone from scratch.

\subsubsection{Proposal Filtering and Linking} Given a new frame of detections $D_1$ and $l-1$ frames of past detections $\{D_2, D_3, \ldots, D_l\}$, an online object tracker is used to stitch the detections of $D_1$ onto pre-existing tubelets $T$ formed by the past frames. This is done by assigning some detections with a track ID $tr_{i\in{n}}$ while discarding others. This turns the list of detections $D_1=[(c_{1, 1}, b_{1, 1}), \ldots, (c_{1,m}, b_{1,m})]$ into tracked detections of $n$ active agents $D'_1=[(c'_{1, 1}, b'_{1, 1}, tr_{1, i\in{n}}), \ldots, (c'_{1,n}, b'_{1,n}, tr_{1, i\in{n}})]$. Detections without an ID can also form a new tubelet should they overlap with other non-tracked detections three frames in a row. 

We adopt OC-SORT's \cite{ocsort2022} tracking scheme to link agent detections into tubes. OC-SORT contains several good attributes that make their algorithm viable for autonomous vehicles. These include:
\begin{enumerate}
\item{
    Fast, online approach to object tracking.
}
\item{
    Robust tracking against occlusions and non-linear paths.
}
\end{enumerate}
In order to eliminate the possibility of different object classes in the same track, we separately track agents from different agent classes.

\subsection{Action Classification with Interaction Encoding and Tube Priors}

\label{section:action_class_w_tube_priors}

\textbf{Encoding Tubelets with DROI-Alignment:} Given agent tube proposals, the action classifier considers sliding windows of $l$ frames as clips. For every clip, it selects the center frame or key-frame (indexed $l/2$) and uses information from the temporal extent of $l$ frames to produce a confidence vector $\{ c_{l/2, j} ^ 1, c_{l/2, j} ^ 2, \dots, c_{l/2, j} ^ n \}$ where $n$ is the number of action classes.

Our approach adapts ACAR-Net \cite{acar2020} to accept the position of a keyframe object $j$ throughout the clip $T_{j} = \{b_{1, j}, \ldots, b_{l, j}\}$ where $j$ must exist in the key-frame and every $b_{i, j}$ has the same track id. To localize an object's features in a specific frame, we modify the use of ROI-align in ACAR-Net. Originally, ACAR-Net performs ROI-alignment across $TemporalAveragePool (F_f)$ and $TemporalAveragePool (F_s)$ using the bounding box $b_{l/2, j}$ of the object present in the key-frame. This design was justified by the original ACAR-Net authors due to the limited movement of agents in the Kinetics Dataset. Particularly in human activity detection in datasets, it can be observed that agents maintain relatively consistent spatial locality throughout frames.

In contrast to the human activity datasets, ROAD agents include vehicles which move dynamically across spatial regions in frames. As a result, an object's track throughout a clip often drastically overreaches the exact bounding box of its position at the key-frame. If, within a clip, an object moves out of its bounding box in the key-frame, then ACAR-Net would no longer be encoding information about the object but unrelated artifacts in the background. This could cause unwanted noise and loss of crucial information. 

 In order to consider Tube Priors, RALACs implements a novel ROI-alignment configuration: Dynamic Region of Interest Alignmnet (DROI). Instead of aligning the key-frame bounding boxes on a temporally pooled feature tensor, we first align the bounding boxes of each frame to its corresponding temporal step in the feature tensor whenever possible and only then pool them. We note that the SlowFast backbone allows us to select feature tensors corresponding to each frame for this purpose.

\textcolor{changes}{Suppose for an input of $l$ frames we have a set of temporally-indexed fast and slow feature tensors $F_f=\{f_{0}, f_{1}, f_{2}, \dots, f_{l-1}\}$ and $F_s=\{s_0, s_{\alpha}, s_{2\alpha}, \dots\}$ from SlowFast. Here, a slow feature exists every $\alpha$ frames where $\alpha$ is the ratio of fast features to slow features. A set of an agent’s temporally-indexed bounding boxes $bb = \{b_0, b_1, b_2, \dots, bb_{l-1}\}$ also exists from the agent tube priors. We apply ROIAlign on each $f_t$ and $s_t$ with the bounding box $b_t$ where $t$ is the temporal frame to extract ROI feature maps $roi_s$ and $roi_f$. We then finally average pool all ROI feature maps along the temporal dimension as shown in Alg. \ref{alg:ped-move}. Since the ROI features are extracted from each frame, the resulting $F_s'$ and $F_f'$ DROI-aligned actor feature maps are therefore encoded with the dynamic bounding box of the agent.}

\begin{algorithm}
    \caption{DROI-Alignment Adapted to SlowFast}
        \begin{algorithmic}[1]
        \color{changes}
        \renewcommand{\algorithmicrequire}{\textbf{Input:}}
        \renewcommand{\algorithmicensure}{\textbf{Output:}}
        \Require Slow feats \textit{$F_s$}, Fast feats \textit{$F_f$}, Bounding boxes \textit{bb}
        \Ensure Feature Tensor with Encoded Tracks through Tube ROI-Alignment
        \State $F_{f,roi} \gets []$ \Comment{ROI-Aligned fast feature frames}
        \State $F_{s,roi} \gets []$ \Comment{ROI-Aligned slow feature frames} 
        \State $\alpha \gets TemporalLen(F_f)/TemporalLen(F_s)$
        \For{$\textit{t, $f_{t}$} \in \textit{TemporalEnumerate($F_f$)}$}
            \State $b \gets bb[t]$ \Comment{Select bounding box at frame}
            \If{$(t + 1) \mod \alpha = 0$} \Comment{If frame has slow feat}
                \State $s_t \gets F_s[t]$ \Comment{Select slow feat at temporal frame}
                \State $roi_s \gets ROIAlign(s_t, b)$ \Comment{Get slow ROI feats}
                \State $F_{s,roi}.append(roi_s)$
            \EndIf
            \State $roi_f \gets ROIAlign(f_t, b)$ \Comment{Get fast ROI feats}
            \State $F_{f,roi}.append(roi_f)$
        \EndFor
        \State $F_f' \gets TemporalAvgPool(F_{f,roi})$
        \State $F_s' \gets TemporalAvgPool(F_{s,roi})$
        \State \Return $F_f', F_s'$
        \end{algorithmic} 
        \label{alg:ped-move}
\end{algorithm}

\textbf{Encoding Road Agent Interactions:} Although the current DROI-Aligned feature maps are sufficient enough to begin classification, our work proposes an extension of ACAR-Net's\cite{acar2020} higher-order relations reasoning to further improve accuracy. Our approach focuses on the hypothesis that actions of a road agent are largely the result of their interactions with other agents and that these interactions can be learned by a neural network.

The idea of capturing higher-order relations between individual human agents interacting with the environment was originally pitched by the authors of ACAR-Net. In their work, the encoded features of human agents, each tiled and concatenated with the features of the overall clip and then reduced using a convolutional layer, were subjected to a modified attention mechanism they called HR$^2$O. To explain this mechanism, given feature maps of $N$ human agents interacting with their environment, $\{F_i\}^N_{i=1}$, their resultant agent-context-agent feature maps $\{H_i\}^N_{i=1}$ are obtained by first converting each $F_i$ into key $K_i$, value $V_i$, and query $Q_i$ embeddings.
\begin{equation}
      Q_i, K_i, V_i = Conv2D(F_i) \\
\end{equation}

An $\tilde{H}_i$ is computed through a linear combination of all other agent value, $V_j$, features according to their computed attention weight, $Att_{i, j}$.
\begin{equation}
       Att^{(x, y)}_{i, j}=softmax_j\left(\frac{\langle Q^{(x, y)}_i , K^{(x, y)}_j\rangle}{\sqrt{d}}\right) \\
\end{equation}
\begin{equation}
        \tilde{H}^{(x, y)}_i=\sum_j Att^{(x, y)}_{i, j}V^{(x, y)}_j \\     
\end{equation}
$\tilde{H}_i$ is then subjected to layer normalization and dropout to become $H_i$.
\begin{equation}
        H_i=Dropout(Conv2D(ReLU(norm(\tilde{H}_i)))) \\
\end{equation}
$H_i$ is then combined with $F_i$ through residual addition to become the final feature map $F'_i$. 
\begin{equation}
       F'_i=F_i + H_i \\
\end{equation}
It is important to note that the attention mechanism is only applied to feature maps along the same spatial location, reducing computation time and limiting HR$^2$O to first-order interactions only. This means that the features of $F_i$ at the spatial location $(x, y)$, denoted as $F^{(x, y)}_i$, are cross-examined only with the other feature maps at the same spatial location, $F^{(x, y)}_j$. 

Once all $\{F'_i\}^N_{i=1}$ are obtained, each agent-context-agent feature map is passed into a fully-connected layer with a non-linearity function to produce the final action confidence scores of each agent.

In ACAR-Net, all feature maps entering their interaction encoding, HR$^2$O, are constrained to human agents only. \textcolor{changes}{In RALACs, we extend its functionality in our Interaction Encoding algorithm to instead populate $\{F_i\}^N_{i=1}$ with the feature maps of multiple different agent classes. Since the feature maps of all agent classes have the same dimension, the HR$^2$O interaction can be used to motivate the model to implicitly learn interactions between all road-relevant classes. During back propagation, the gradient for a specific agent's feature map will be derived from all other DROI-aligned agent feature maps, along with the environment feature map. This enables robust action classification of multi-class agents in complex scenes. }

\color{changes}
Given the produced set of confidence vectors for N agents $X = \{C_0, C_1, \dots, C_{N-1}\}$, where $C_j$ contains the confidence of n actions for the $j$th actor $C_j = \{ c_j^1, c_j^2, \dots, c_j^n \}$. We apply a sigmoid cross-entropy focal loss \cite{8417976} to the predicted action confidence vector against the ground truth action vector $Y_j=\{y_j^1, y_j^2, \dots, y_j^n\}$ for training.
\begin{equation}
    \mathcal{L}_f(p_t) = -\alpha(1 - p_t)^{\gamma}\log{\left(p_t\right)}
\end{equation}
\begin{equation}
    g(p, y) = \begin{cases}
        p & \text{if } y = 1\\
        1 - p & \text{otherwise.}
    \end{cases}
\end{equation}
\begin{equation}
    \mathcal{L}_{action} = \sum_{j=0}^{N-1}{\sum_{i=1}^{n}{\mathcal{L}_f\left(g\left(c_j^i, y_j^i\right)\right)}}
\end{equation}

The focal loss helps overcome the severe class imbalance of the action class distributions in the ROAD Dataset, in which there is a larger majority of vehicle and pedestrian classes present in the training data as opposed to cyclists, emergency vehicles, etc.
\color{black}

\begin{figure*}[h!]
    \centering
  \subfloat[\centering \textbf{Optical Flow:} Ego Vehicle Moving Right. Flow pattern differs at truck and cyclist, but not parked cars]{%
       \includegraphics[width=0.33\linewidth]{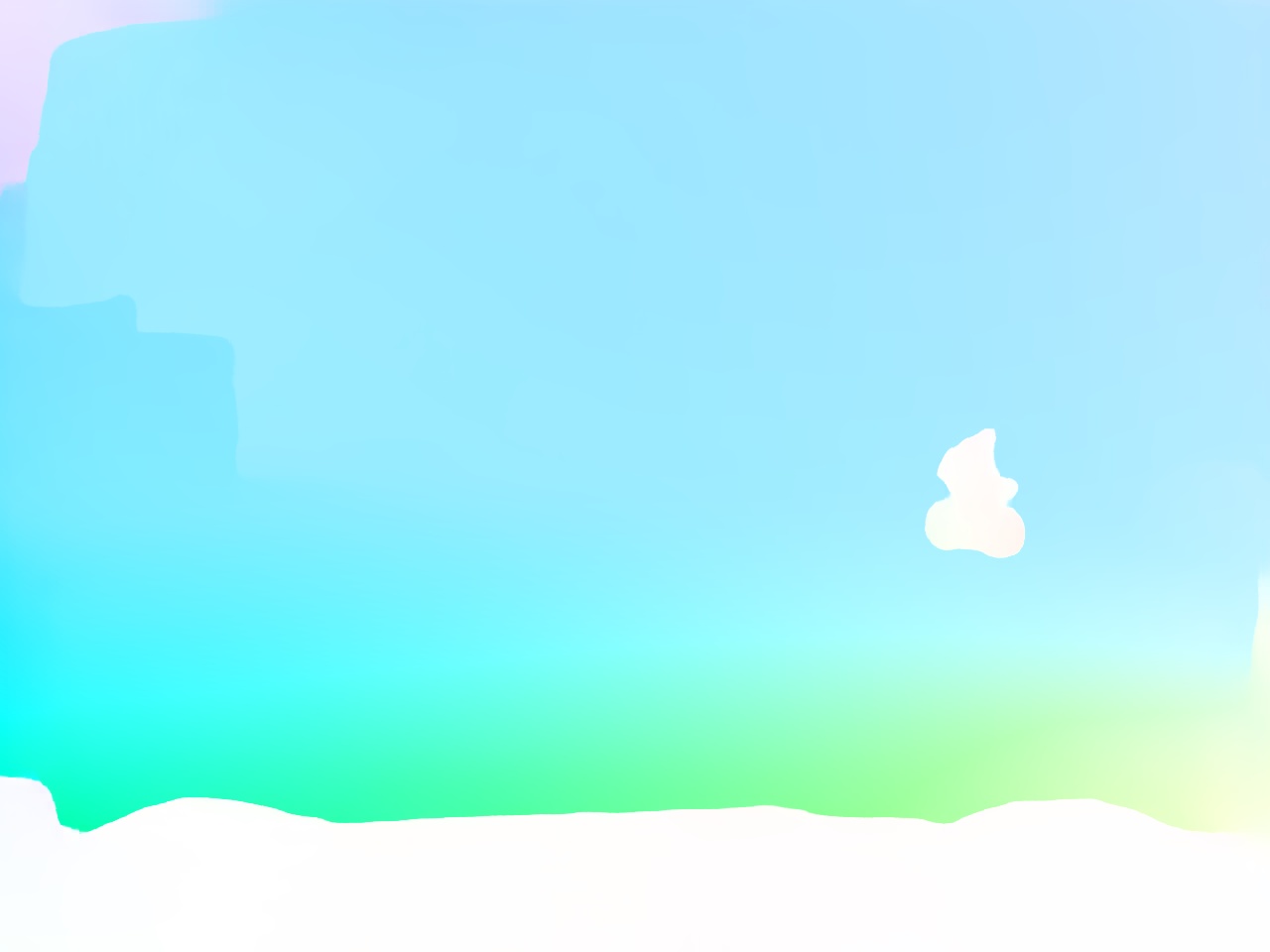}}
    \hfill
  \subfloat[\centering \textbf{Incorrect:} Action Detector falsely detects parked cars]{%
        \includegraphics[width=0.33\linewidth]{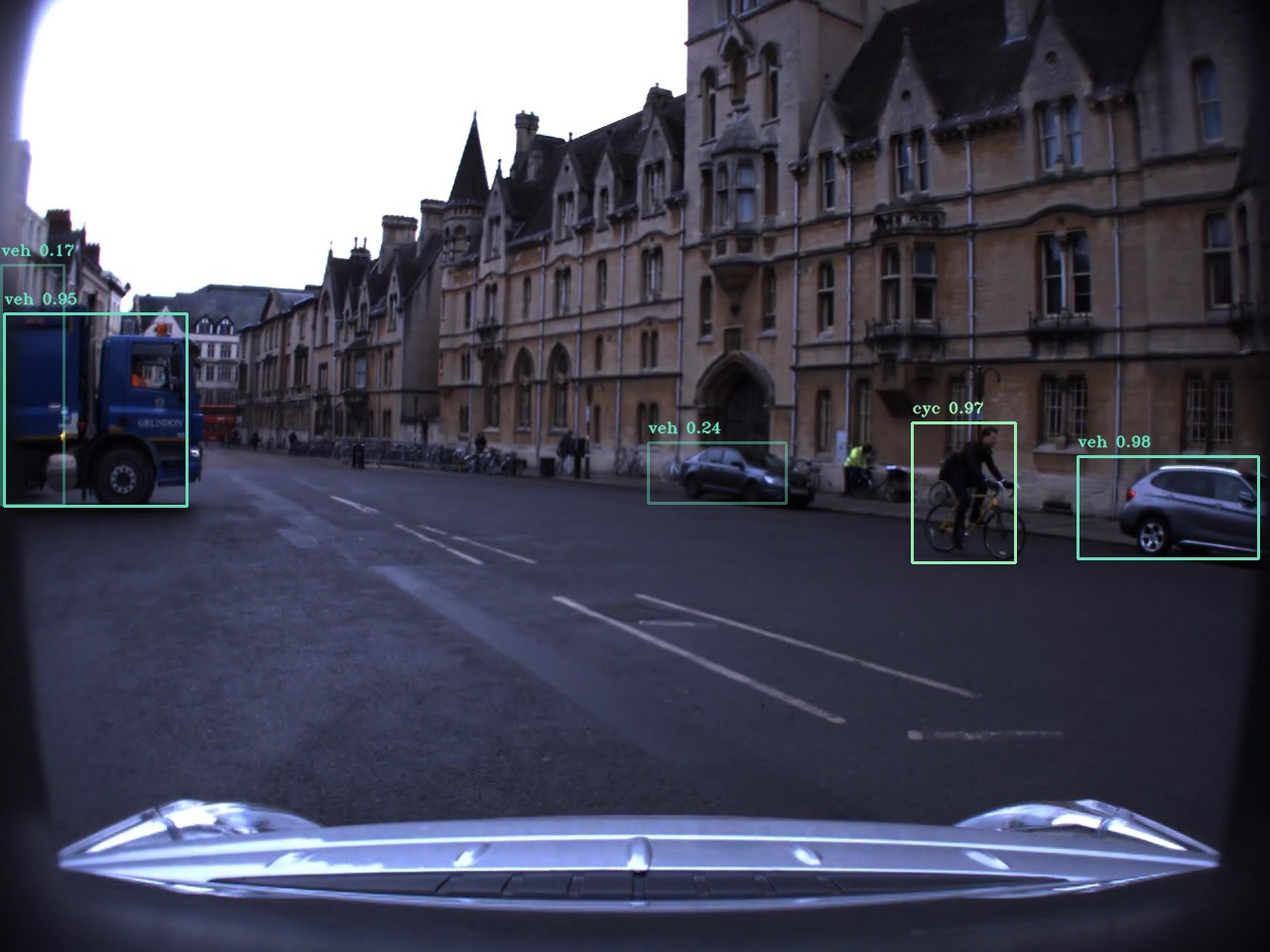}}
    \hfill
  \subfloat[\centering \textbf{Correct:} Flow-Based Action Detector takes into account unchanging flow pattern around parked vehicles to recognize them as inactive]{%
        \includegraphics[width=0.33\linewidth]{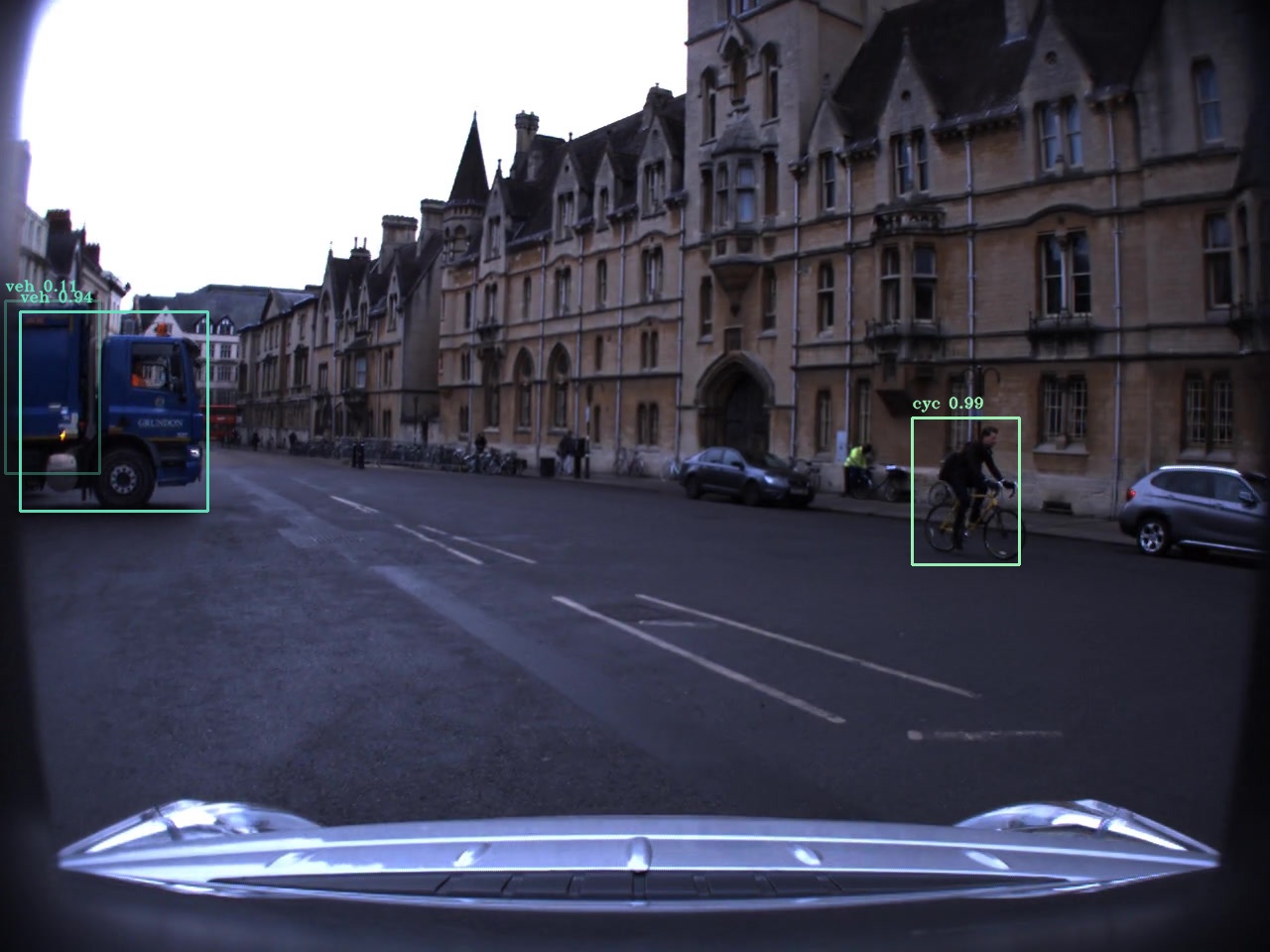}}
  \caption{Action Localization comparison between RGB-only and Flow-fused detectors. The use of optical flow provides the network sufficient enough information to classify an agent as active or inactive.}
  \label{fig:action:optical}
\end{figure*}

\section{Experimentation}
\label{section:experimentation}

\subsection{ICCV2021 ROAD Challenge}
We focus our performance evaluations and ablation studies using the ICCV2021 ROAD Dataset \cite{singh2022road}. Being the first dataset of its kind, the ROAD Dataset consists of 22 videos of annotated road actions on a moving ego vehicle. These videos are each 8 minutes long and recorded at an average of 12fps. The result is over 120K frames with 640K annotated actions spread throughout. ROAD no longer accepts test set submissions, so we base our experimental validation solely on its publicly available training and validation data. 

The ROAD dataset is organized into 3 splits, each of which represents different sets of validation scenarios. Splits 1, 2, and 3 validate on overcast, nighttime, and sunny scenarios respectively. 

The ROAD dataset provides us with thorough road-relevant action labels which form the basis of all experimental analysis in this paper. However, its use of action tracks for official evaluation pushes for an action-centric design which is quite unorthodox. Because environmental modeling on autonomous vehicles have traditionally relied on object tracks, our suggestion is to maintain the integrity of object tracks and have its frame-wise action appended as an attribute instead. This approach not only enables the ease of integration on existing autonomy stacks, but also provides room for future attribute expansion. For example, a car track could be appended with attributes like whether or not the car is following road rules, or perhaps a metric for the car user's mood or historical tendency to drive recklessly. Nevertheless, we note that the use of action tracks for evaluation is limited and we only use this metric to compare with the baseline model. All other experimental metrics are done at the frame level.  

\subsubsection{Action Track Post-processing}
To compare our results to the existing baseline and evaluate our end-to-end performance using the ROAD dataset's official evaluation scheme, our frame-wise action detections must be processed to form action tracks. We perform tube trimming to process the final actions into tube proposals that cater to the official ROAD dataset evaluation. This can be implemented as an offline or online approach.

\textbf{Tube Trimming:} 
We solve the problem of action track specificity by trimming agent tubes that already exist. Suppose we have frames in an agent's tube $T = \{f_{ik}, \ldots\}$ with confidence scores $s_{ik}$ for class $k$. Intuitively, we would like to partition $T$ into ordered subsets $T = \{F_a, F_b, F_c\}$ where for some threshold $\varepsilon$, $s_{ik} < \varepsilon$ when $s_{ik} \in F_a \cup F_c$. This operation would isolate the action in $F_b$ and remove noise and other actions belonging to the same agent in $F_a \cup F_c$. We note, however, that this is not possible in online contexts, so for these cases we simply apply a mask over the frame scores based on the parameter $\varepsilon$, typically set to a value $\leq 0.001$.

\subsubsection{Implementation Details}
\label{section:implementation}
\textbf{Action Localization:} We use Faster R-CNN with a modified Feature Pyramid Network \cite{lin2017feature} neck and ResNeXT-101 backbone as the object detector. We pre-trained the model on COCO with a single RTX 3090 and duplicate its backbone for both optical flow and RGB channels. The backbones of both channels are jointly trained with the detection head. We train the detection model end-to-end for 7 epochs ($\sim$140k steps) with a batch size of 16. To stitch these detections into tracks, OCSORT with an IOU of 0.3, and Kalman filter parameters $\delta t = 3$ and $inertia = 0.2$, were used. 

\textbf{\textit{Inactive Agent Pseudo Annotations}}
The ROAD dataset only annotates active agents (i.e., a parked on the side of the road is not considered to be ’active’, but a stopped vehicle at an intersection is), and the lack of explicit annotations for inactive agents generates noisy learning targets for the object detector which could degrade performance. To solve this problem, we propose to generate pseudo-labels for the inactive agents on the ROAD dataset. Since most of the inactive agents are vehicles, we only consider them in our method. 
First, we train the Faster R-CNN FPN \cite{lin2017feature} with ResNeXt-101 as backbone on the COCO dataset \cite{lin2014microsoft}, and we use it to predict vehicles in the ROAD dataset. After obtaining the vehicle detections, we filter for detections which have a confidence score of 0.9 or higher and an intersection over union of less than 0.2 with the ground truth ROAD detections. These detections are then relabeled as a new 'inactive agent' class. We use these detections, along with the original annotations, to train our final active agent detector. We found that this approach yields better results than training our detector directly on the data.  

\textbf{Action Classification:} We use officially released SlowFast weights pre-trained on the Kinetics-400 dataset \color{changes}\cite{kay2017kinetics}\color{black}. We allow these weights to be fine-tuned during training, and do not use any pre-trained weights for the interaction encoding block. We observe our best results on average after training for $\sim$140k steps or $7-9$ epochs on dual RTX 3090s with a batch size of 8. We use a sigmoid cross-entropy focal loss \textcolor{changes}{with $\alpha=0.25$ and $\gamma = 2$}  to overcome the severe class imbalance of the action class distributions in the ROAD Dataset. We use a base learning rate of 8e-4, which decreases by a factor of 10 by epochs 4, 6, and 7. We perform linear warm-up during the first epoch at a learning rate of 6.4e-2, and use a weight decay of 1e-5 with Nesterov momentum of 0.9.

\textcolor{changes}{The full end-to-end model has 243M parameters and inferences at 15 frames per second on an RTX 3090.}

\subsection{Full Model Evaluation}
We evaluate our proposed method against the current baseline model, 3D-RetinaNet, as it is the only model with comprehensive validation metrics. Table \ref{tab:iccv_road_results} shows our results on the ROAD dataset. We see that RALACs' online performance outperforms the current baseline.

\begin{table}[h]
\begin{center}
\normalsize
\begin{tabular}{ l | c | c | c | c}
\hline
Frame-Level Eval & Val 1 & Val 2 & Val 3 & Avg \\
\hline
3D-RetinaNet (frame-level) & 26.2  & 11.7& 21.2  & 19.7\\
\textbf{Ours (frame-level)}& \textbf{31.9} & \textbf{12.4} & \textbf{22.9} & \textbf{22.4}\\
\hline 
Video-Level Eval & Val 1 & Val 2 & Val 3 & Avg \\
\hline
3D-RetinaNet & 17.0  & 11.4 & 14.6 & 14.3\\
\textbf{Ours (Online)}  & \textbf{22.7} & \textbf{11.5} & \textbf{15.4} & \textbf{16.5} \\
Ours (Offline) & 23.6 & 11.7 & 16.1 &  17.1 \\
\hline
\end{tabular}
\vspace{1mm}
\caption{Frame and video-mAP scores on the ROAD Dataset. We compare frame-mAP@0.5IOU and video-level mAP@0.2IOU following the 3D-RetinaNet baseline. Experimentally, it is observed that these results may fluctuate by $\sim$1\%. For our online results, Tube-ACAR considers a temporal extent of 32 frames, so we are able to perform box interpolation within gaps of 32 frames or less.}
\label{tab:iccv_road_results}
\end{center}
\end{table}

Breaking down the evaluation, we see that our overall approach functions considerably better in ideal overcast scenarios. Val 1 is characterized by little weather-related faults such as rain and glare from vehicles and the sun. In contrast, Val 2 and Val 3 have a considerable number of these faults, and our performance consequently has room for improvement. We also note that Split 2 does not have any training data at night, so Val 2 was our system's first exposure to nighttime scenarios. 

Nevertheless, our overall performance outperforms the current baseline despite these pitfalls.

\definecolor{wato_grey}{HTML}{555555}
\definecolor{wato_blue}{HTML}{DAE8FC}
\definecolor{wato_yellow}{HTML}{FFF2CC}
\begin{figure*}
    \centering
    \begin{tikzpicture}
        \begin{axis}[
            ybar,
            enlargelimits=0.08,
            legend style={at={(0.5,-0.25)},
            axis x line*=none,
            axis y line*=none,
            anchor=north,legend columns=-1},
            ylabel={AP@0.5IOU},
            symbolic x coords={Overall,Braking,TurLft,TurRht,XingLft,XingRht,Wait2X,Mov,MovAway,MovTow},
            xtick=data,
            nodes near coords,
            nodes near coords align={anchor=west},
            nodes near coords style={rotate=90},
            width=18cm,
            height=5cm,
        ]
        \addplot [fill=wato_grey]coordinates {(Overall,0.34) (Braking,0.09) (TurLft,0.14) (TurRht,0.12) (XingLft,0.30) (XingRht,0.28) (Wait2X,0.35) (Mov,0.58) (MovAway,0.65) (MovTow,0.71)};
        \addplot [fill=wato_blue]coordinates {(Overall,0.37) (Braking,0.26) (TurLft,0.18) (TurRht,0.17) (XingLft,0.35) (XingRht,0.33) (Wait2X,0.51) (Mov,0.56) (MovAway,0.70) (MovTow,0.77)};
        \addplot [fill=wato_yellow]coordinates {(Overall,0.41) (Braking,0.23) (TurLft,0.23) (TurRht,0.26) (XingLft,0.51) (XingRht,0.43) (Wait2X,0.61) (Mov,0.62) (MovAway,0.81) (MovTow,0.86)};
        \legend{SlowFast,SlowFast+Interaction Encoding,SlowFast+Interaction Encoding+DROI Alignment}
        \end{axis}
    \end{tikzpicture}
    \caption{Analysis of Action Classification improvements by action class. We see that the use of Interaction Encoding improves the classification of actions that naturally require more interactions with other agents (see \textit{braking} and \textit{crossing (X)} actions). Furthermore, DROI alignment is also shown to be an effective method for extracting the information present in agent tracks, especially during actions of large movements.}
    \label{fig:classification_ablation}
\end{figure*}
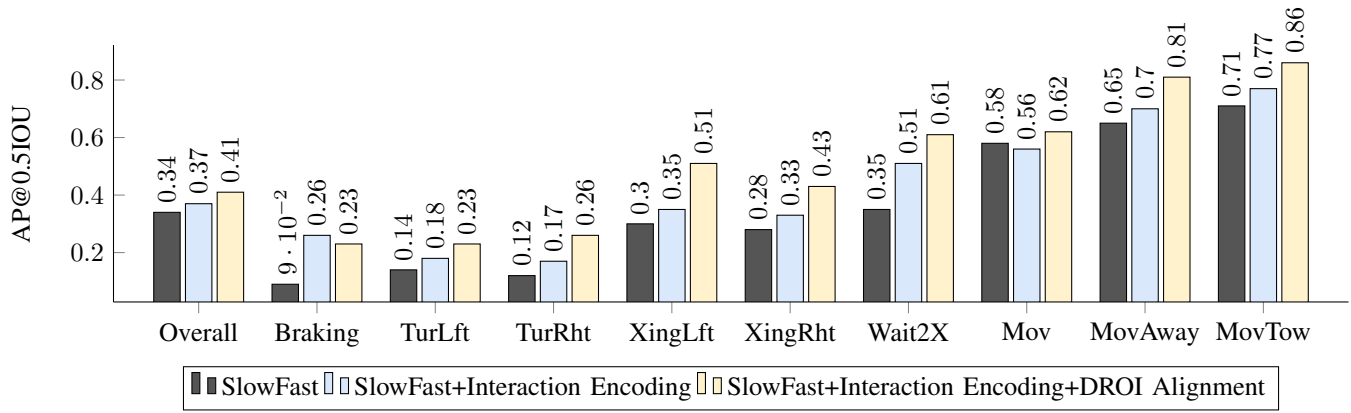

\subsection{Ablation Studies}
\label{section:ablation}
To provide results on the effectiveness of each of our proposed contributions, we perform experiments on each stage of the system in isolation. We use the Val 1 split to best illustrate the effects of our improvements, but a similar trend is apparent in all the other splits. The results are summarized in Table \ref{tab:ablation}. For our Action Classifier, we use ground truth bounding boxes as input and compare its results with ground truth actions provided by the ROAD dataset.

\begin{table}[h]
\begin{center}
\normalsize
\begin{tabular}{ p{7.4cm}| p{0.6cm} }
\hline 
\hspace{30mm}Stage & mAP \\
\hline
Faster R-CNN (Raw) & 67.9\\
Optical Flow + RGB Input Fused Faster R-CNN & 75.4 \\
Faster R-CNN (Pseudo Annotations) & 80.5 \\
\textbf{Optical Flow + RGB FPN Fused Faster R-CNN} & \textbf{83.1} \\
\hline
SlowFast & 34.5 \\
SlowFast + Interaction Encoding & 37.1 \\
\textbf{SlowFast + Interaction Encoding + DROI Alignment} & \textbf{41.1}\\
\hline
SlowFast + Interaction Encoding + FPN Fused Faster R-CNN & 27.8 \\
\textbf{Full Action Classifier + Full Action Localization} & \textbf{31.9}\\
\hline
\end{tabular}
\vspace{1mm}
\caption{Improvements from various techniques explored in the architecture. Detector results are reported in frame-level agent mAP, Action Classification results are reported in frame-level action mAP. Isolated Action Classification experiments observe higher results due to ground truth box inputs. All ablations are reported on validation split 1 since it has the most even class distribution.}
\label{tab:ablation}
\end{center}
\end{table}

\begin{figure}[h!]
    \centering
  \subfloat[\centering \textbf{Incorrect:} Moving Away on Red Light (left)]{%
       \includegraphics[width=0.33\linewidth]{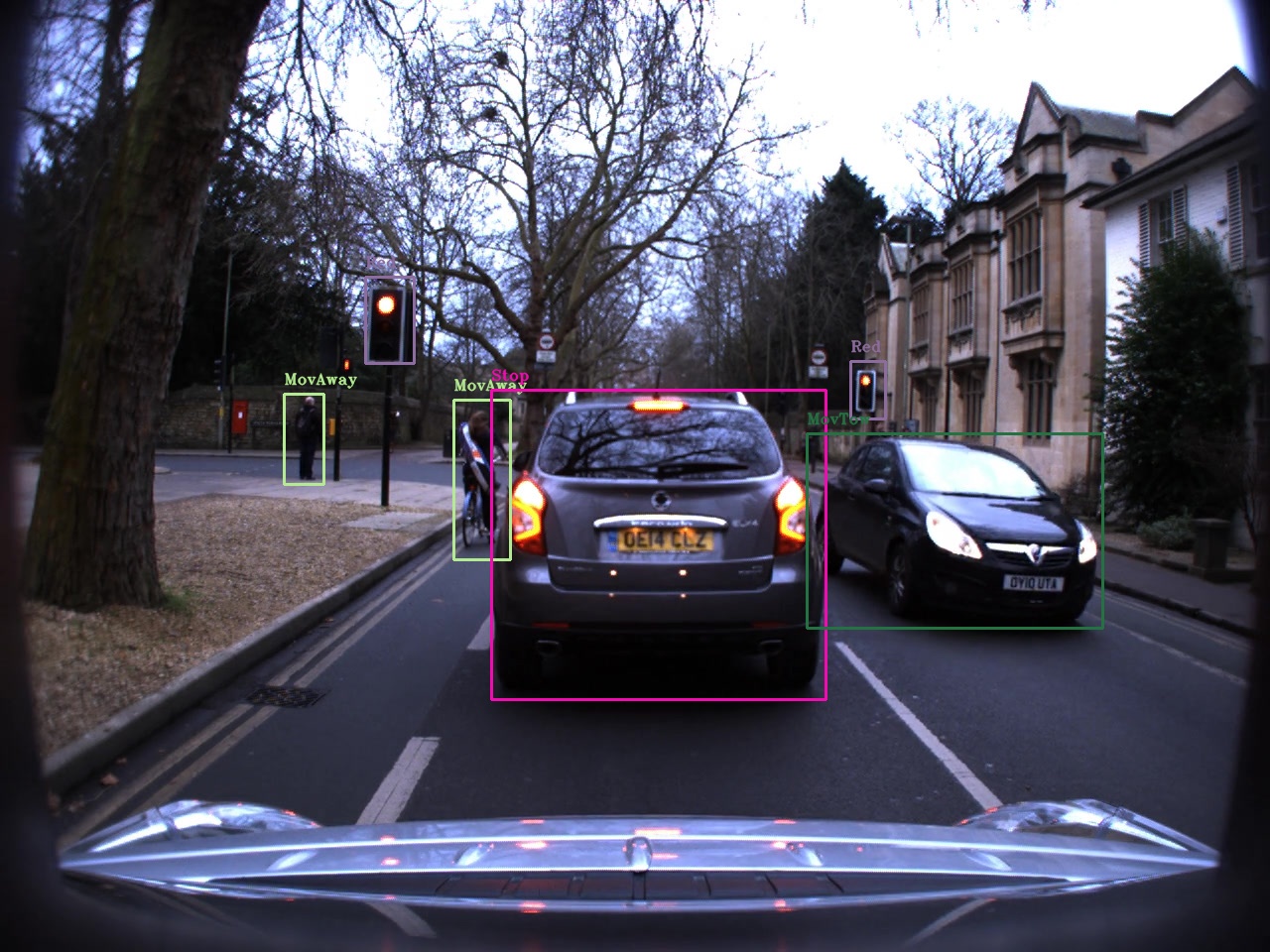}}
    \hfill
  \subfloat[\centering \textbf{Incorrect:} Stopped on Green, Moving without right of way]{%
        \includegraphics[width=0.33\linewidth]{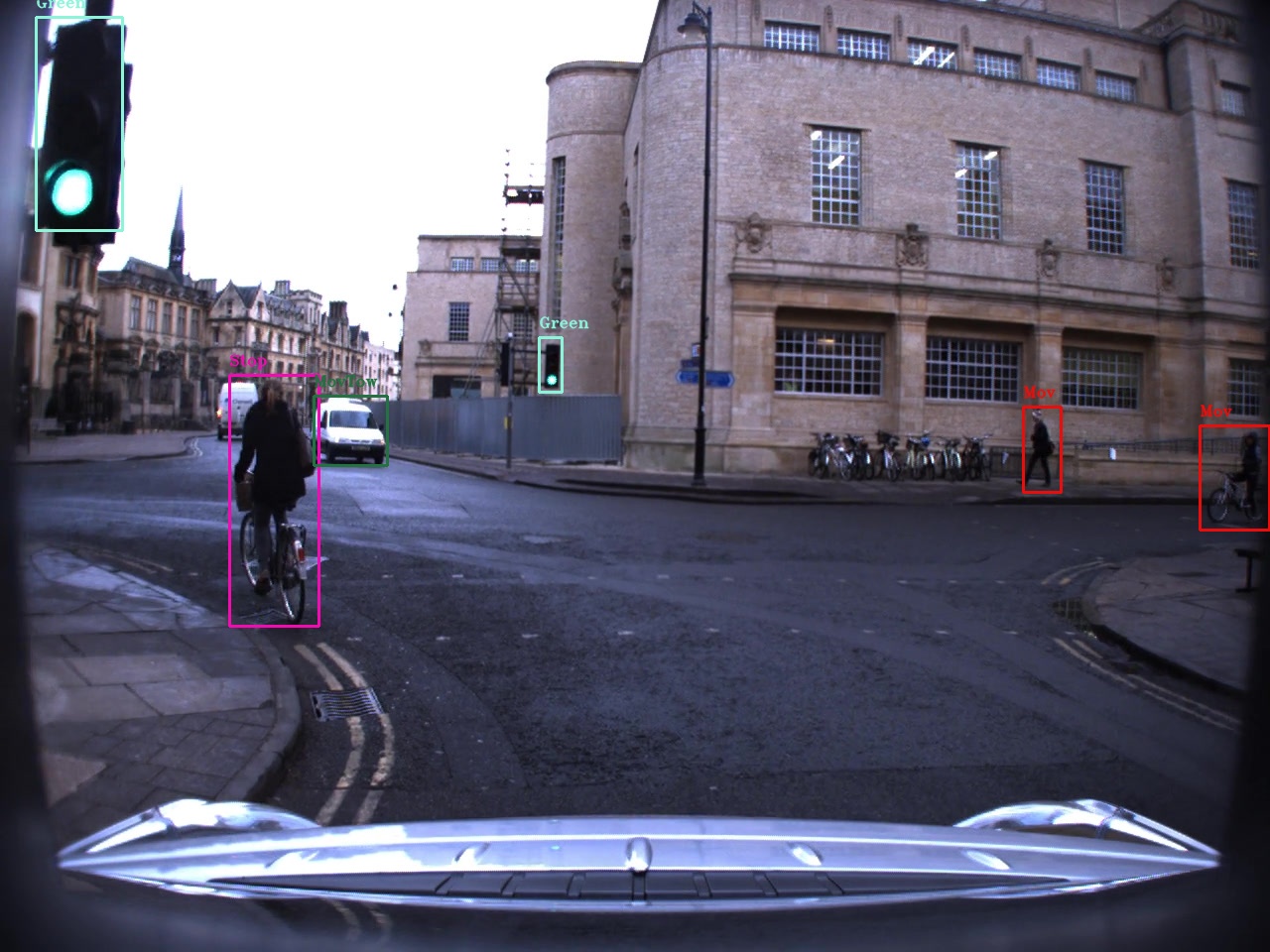}}
    \hfill
  \subfloat[\centering \textbf{Incorrect:} Stopped at Intersection Crossing (left)]{%
        \includegraphics[width=0.33\linewidth]{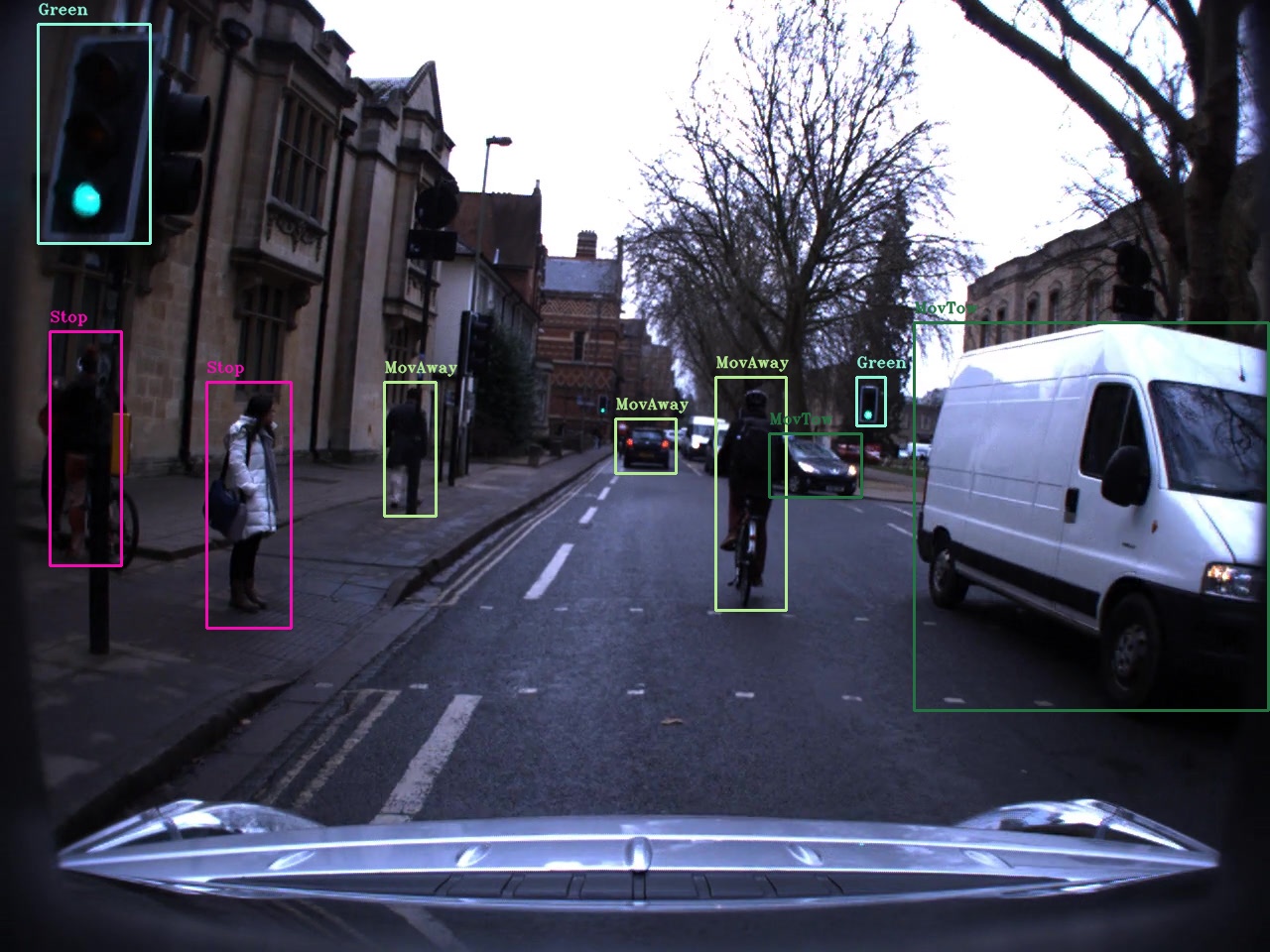}}
    \\
  \subfloat[\centering \textbf{Correct:} Stopped on Red]{%
        \includegraphics[width=0.33\linewidth]{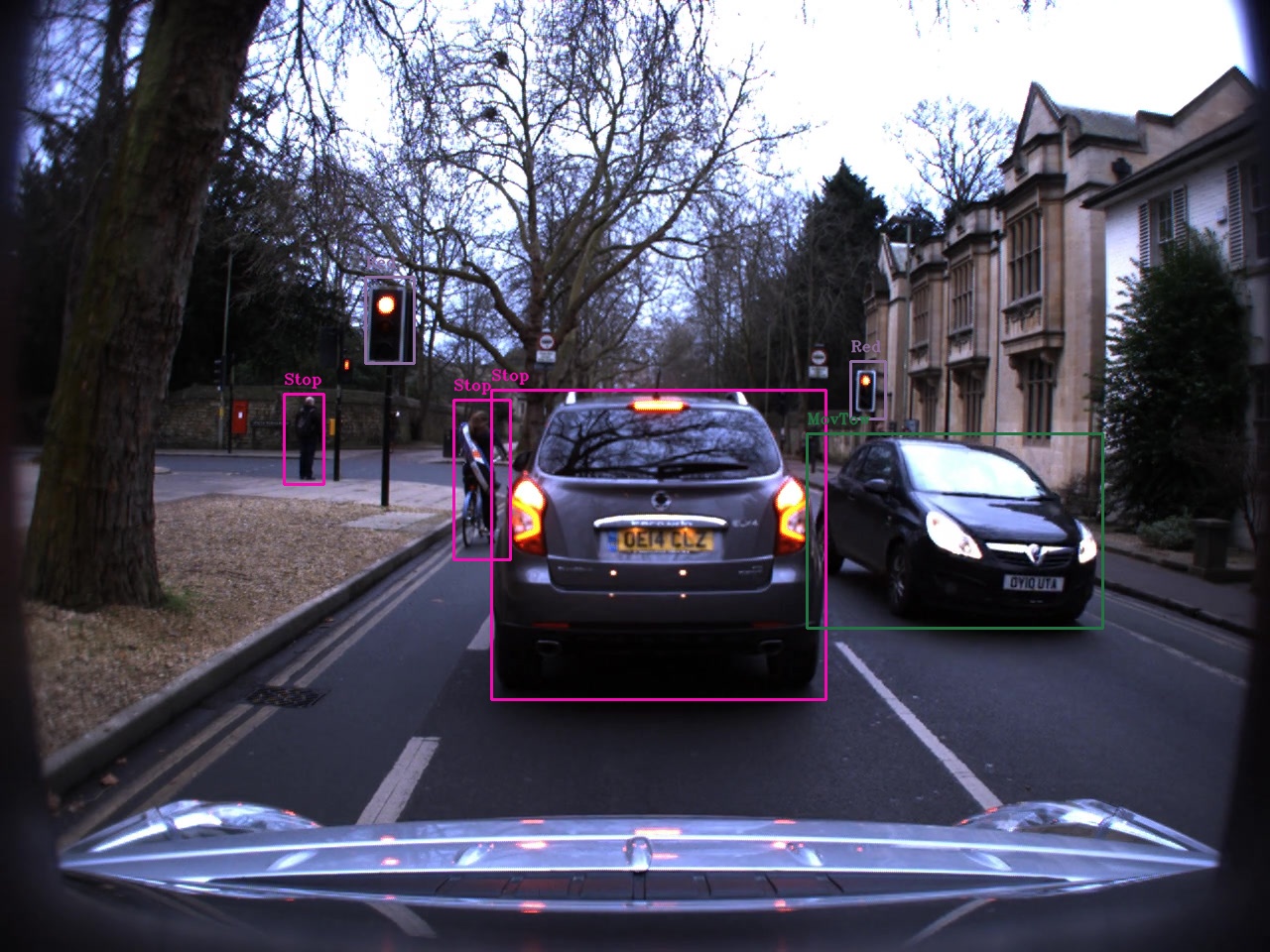}}
    \hfill
  \subfloat[\centering \textbf{Correct:} Moving on Green, Stopped without right of way]{%
        \includegraphics[width=0.33\linewidth]{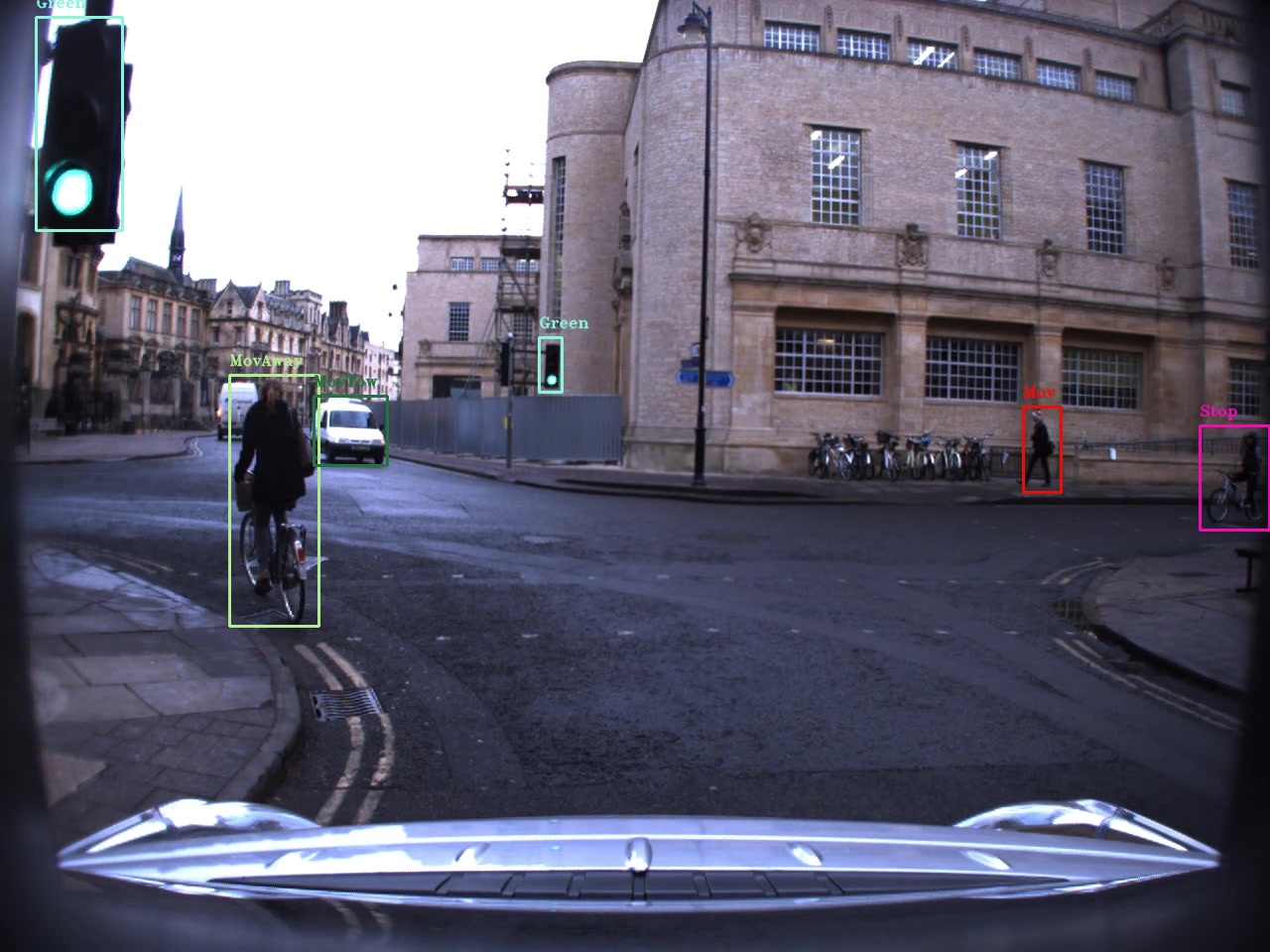}}
    \hfill
  \subfloat[\centering \textbf{Correct:} Waiting to Cross Intersection]{%
        \includegraphics[width=0.33\linewidth]{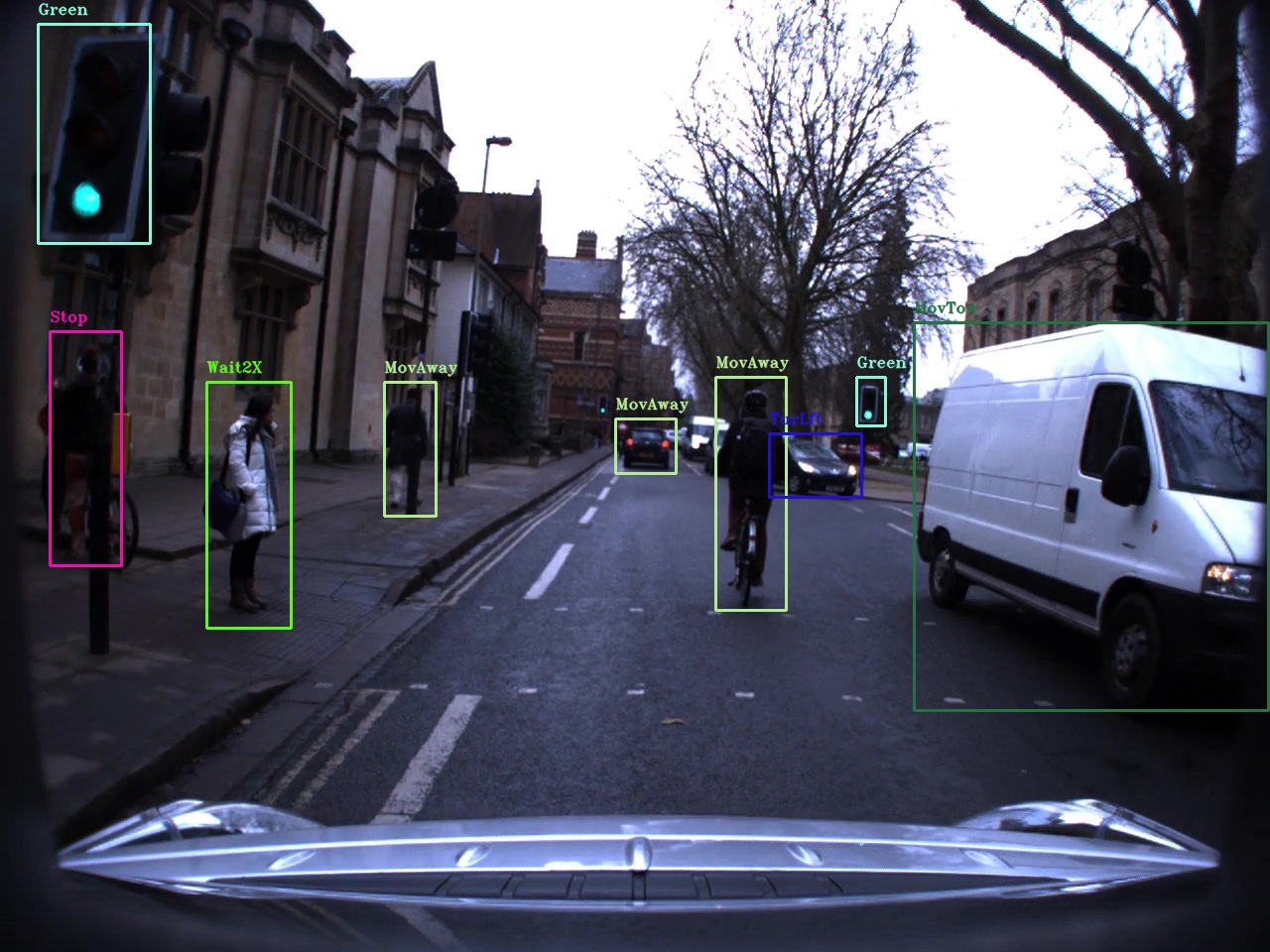}}
  \caption{Action Classification comparison at various intersection scenes. The top row photos (a-c) are actions classified without the use of Interaction Encoding. Consequently, we see that the network seems oblivious to the context of other agents around it (other agents moving on Green, light colours). In contrast, the use of Interaction Encoding (d-f) seems to correct these faults. We attribute this to the network's newfound ability to comprehend action interactions. Note: Only pedestrians can be considered as Waiting to Cross.}
  \label{fig:action:interaction} 
\end{figure}

\textbf{Action Localization Ablation:} In our experiments, we examine two different approaches to fusing optical flow into the action detection problem. Note we use the 3-channel color wheel representation of Optical Flow to retain dimensionality.

\begin{enumerate}
\item{
    \textit{Input Fusion:} The backbone accepts a concatenation of the RGB and the Optical Flow frame.
}
\item{
    \textit{FPN Fusion:} The RGB and Optical Flow frames are each passed through their own backbone and summed together at different scales in the FPN.
}
\end{enumerate}

From Table \ref{tab:ablation}, we see that Input Fusion yields underwhelming results compared to a Faster RCNN trained on pseudo annotations. This could be the result of a multitude of problems, one of which could be the lack of a pre-trained backbone. 

\begin{figure}[h!]
    \centering
  \subfloat[\centering First frame in frame-based input]{%
       \includegraphics[width=0.33\linewidth]{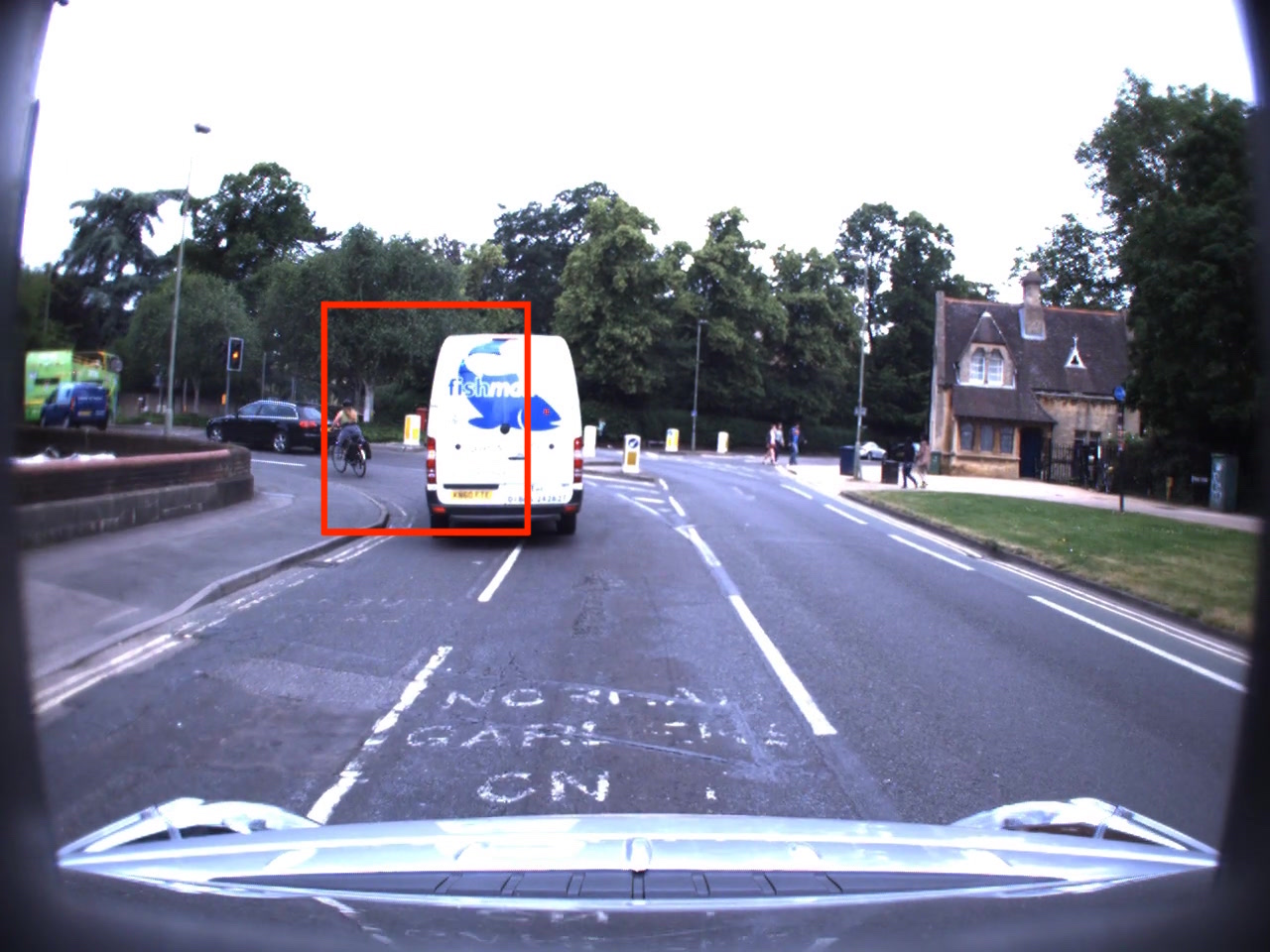}}
    \hfill
  \subfloat[\centering Middle (keyframe) frame in frame-based input]{%
        \includegraphics[width=0.33\linewidth]{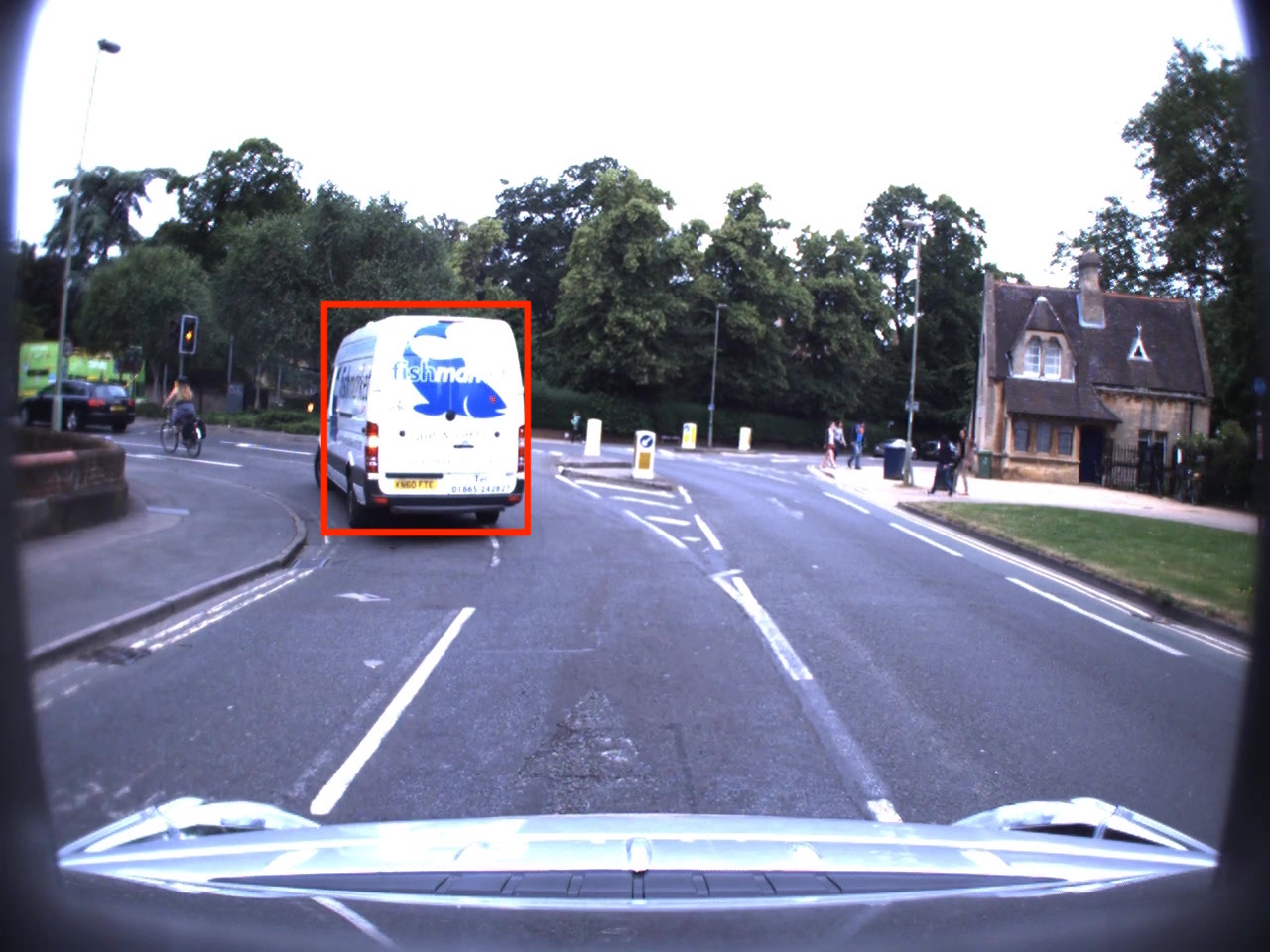}}
    \hfill
  \subfloat[\centering Last frame in frame-based input]{%
        \includegraphics[width=0.33\linewidth]{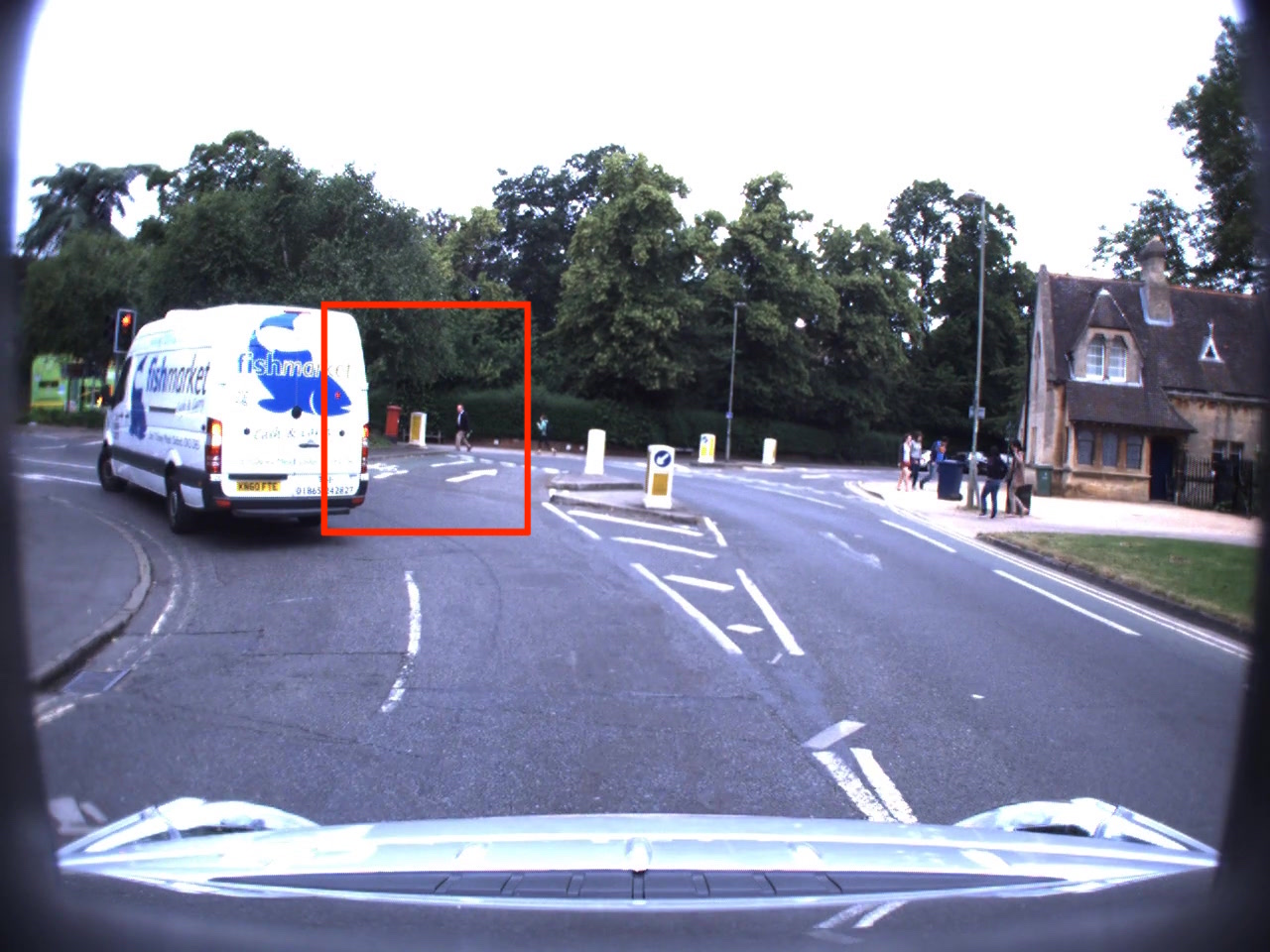}}
    \\
  \subfloat[\centering First frame in tube-based input]{%
        \includegraphics[width=0.33\linewidth]{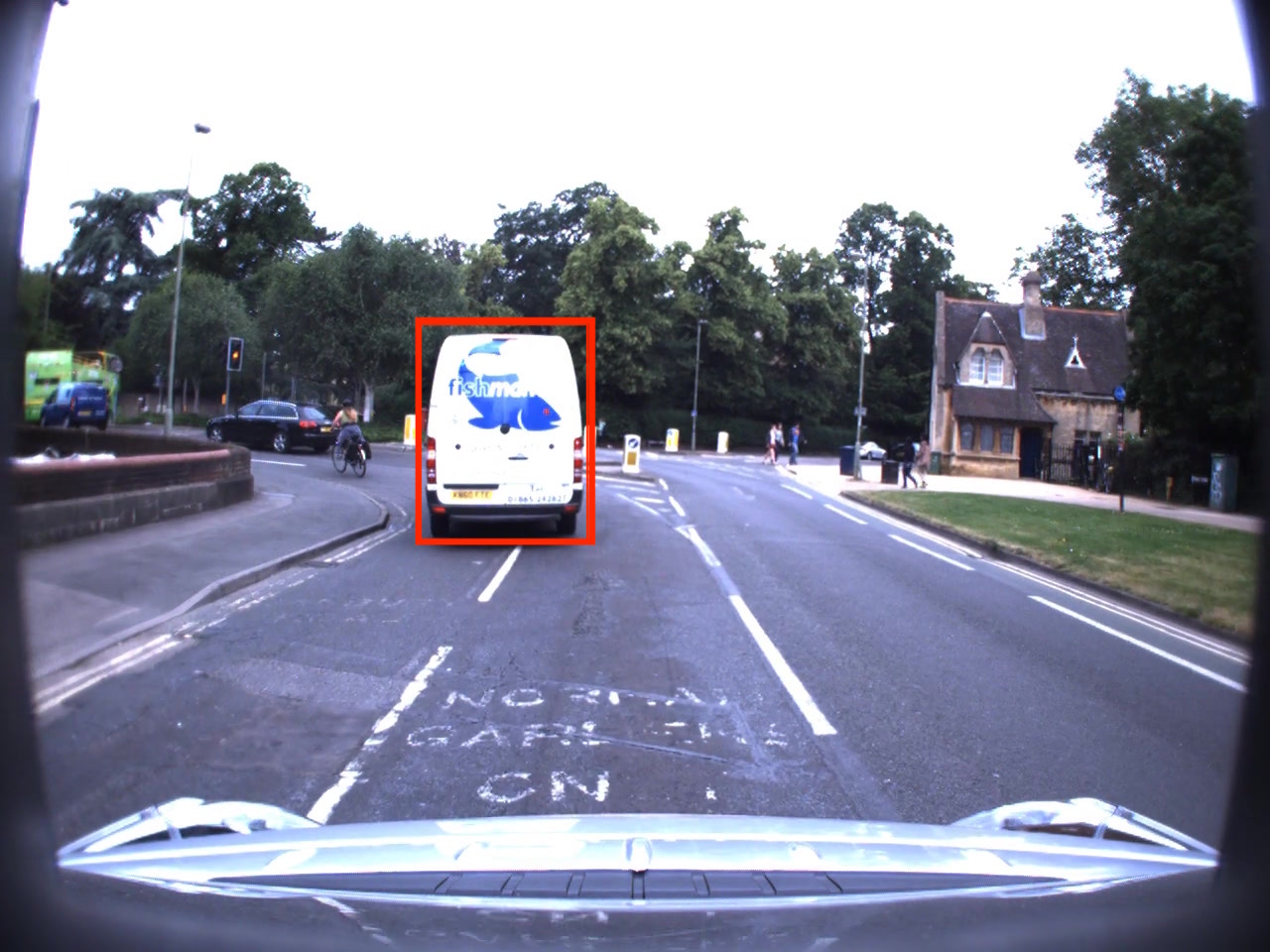}}
    \hfill
  \subfloat[\centering Middle (keyframe) frame in tube-based input]{%
        \includegraphics[width=0.33\linewidth]{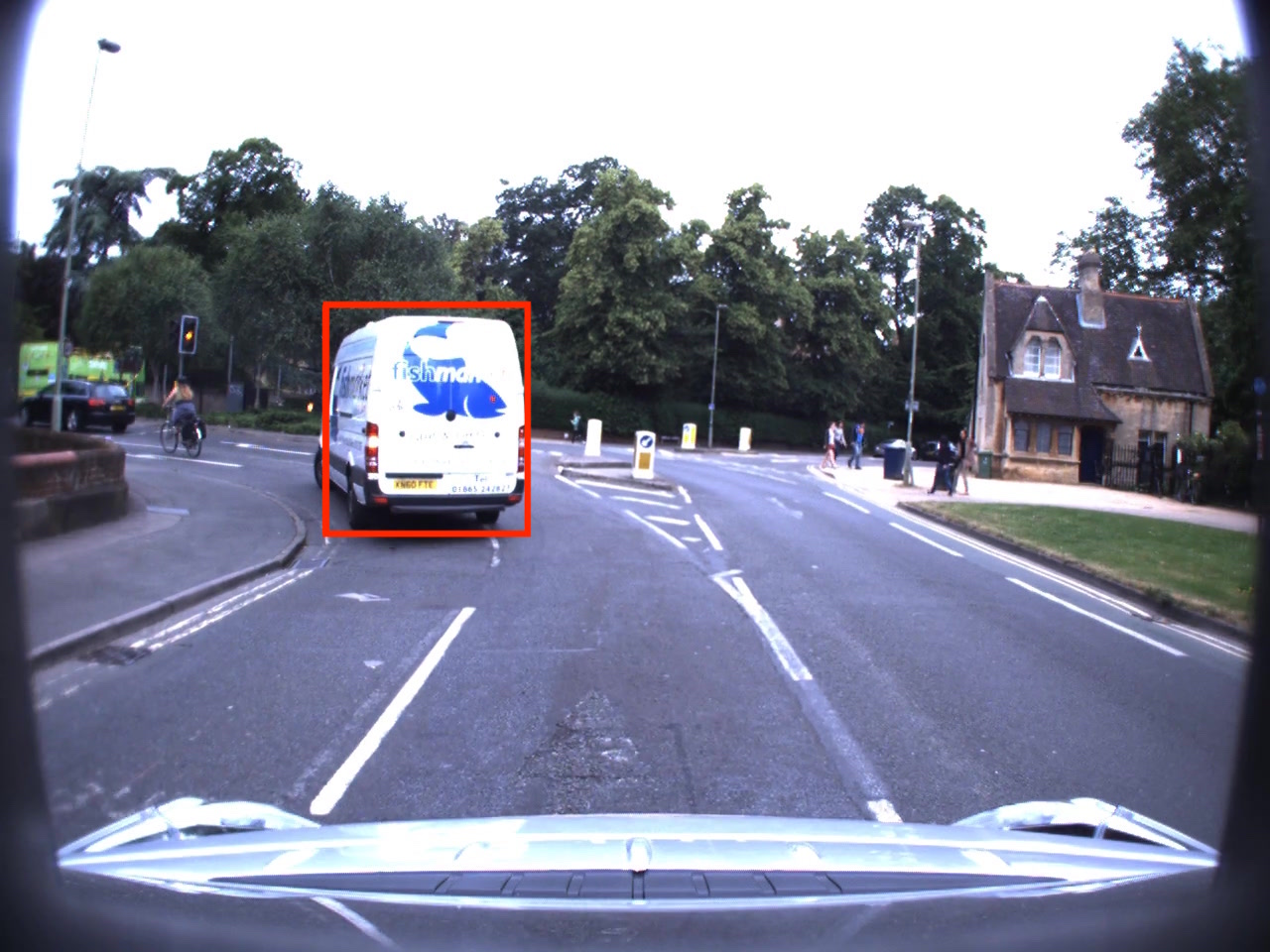}}
    \hfill
  \subfloat[\centering Last frame in tube-based input]{%
        \includegraphics[width=0.33\linewidth]{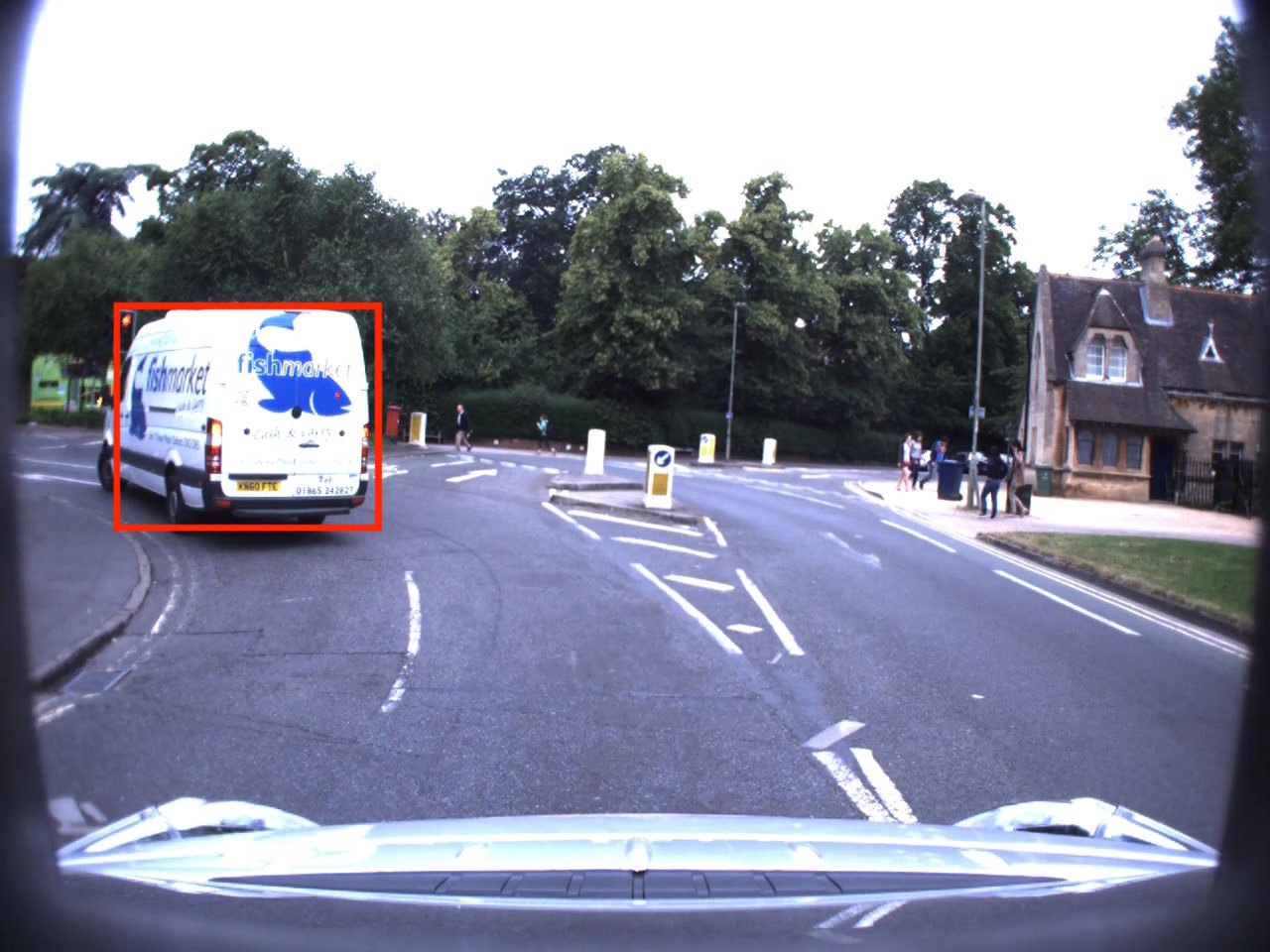}}
  \caption{Difference between inputs to ROI and DROI based action classifiers. The top row (a-c) is the ROI-based input, obtained from just the detector. The keyframe detection (b) does not provide good spatial localization of the agent across the entire temporal view (a) and (b). In contrast, the bottom row (d-f) is the DROI-based input based on the detector and the OC-Sort tracking algorithm. In combination with the other modifications made to the model, the tube provides good spatial tracking across the view, improving action classification accuracy as presented in Table \ref{tab:ablation} and Figure \ref{fig:classification_ablation}}
  \label{fig:action:droi_alignment} 
\end{figure}

In contrast, when Optical Flow is encoded and fused at multiple feature scales in the FPN, we get better results than the RGB-only approaches. Figure \ref{fig:action:optical} provides an example of this. We hypothesize that initially encoding the optical flow and RGB channels implicitly instructs the network to first localize objects and their respective flow field. These localizations are then cross-compared at the summation joint to determine its active/inactive state.

Our results also show that penalizing a pre-trained detector for localizing a non-active agent is not as effective as penalizing it for misclassification. By retraining the detector to detect agents as "inactive" instead of not detecting them at all, we observe faster convergence. 

\textbf{Action Classification Ablation:} To examine the impact of our contributions to action classification for road agents, an in-depth, class-wise ablation analysis was conducted. Figure \ref{fig:classification_ablation} shows various performance metrics of different road action classes. The differences in average precision between classes correlate directly with the frequency of each class in the training data.

\begin{figure*}[h!]
    \centering
    \includegraphics[width=0.8\textwidth]{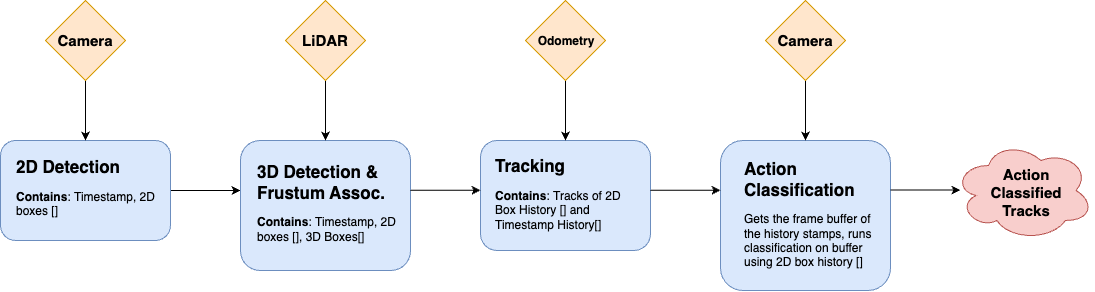}
  \caption{Data pipeline used to deploy RALACs to the perception stack. The resulting action classified tracks are sent to the ADS environment model \cite{dempster2022drg} in which they are used as input to behavioral planning decisions.}
  \label{fig:action:deployment_pipeline}
\end{figure*}

\textbf{\textit{Interaction Encoding:}} Naturally, certain actions depend more on cross-agent interactions than others. In Figure \ref{fig:classification_ablation}, we see a very similar pattern, where actions like \emph{Braking} and \emph{Waiting to Cross (Wait2X)} both benefit significantly from the addition of Interaction Encoding. Both of these actions heavily rely on interactions; as Braking requires some other agent to brake for, and agents Waiting to Cross are waiting for other agents to cross. 

Furthermore, we see that Interaction Encoding also provides a general benefit to the precision of all action classes. We attribute this to the model's increased awareness of action context. That is, the model can now consider the actions of an agent in context with the actions happening around it. This is especially true in interaction-heavy scenes like intersections, as shown in Figure \ref{fig:action:interaction}. 

Most importantly, this overall improvement using Interaction Encoding shows how our network not just successfully learned interactions but interactions that are agent-agnostic. This finding alone shows promise for future interaction encoding schemes, and a possible avenue for future research. 

\textbf{\textit{DROI Alignment:}} Based on our experiments, the addition of DROI Alignment provides an overall benefit to all action classes. We attribute this improvement to the successful encoding of object tracks. Figure \ref{fig:action:droi_alignment} shows a comparison between the data encoded by traditional ROI Alignment as opposed to our DROI strategy. \textcolor{changes}{The traditional frame-based input ROI Alignment strategy uses a static bounding box from the middle keyframe, and does not take into account the dynamic spatial movement of the agent between frames. Our proposed tube-based input encodes the dynamic bounding box between frames into the feature map based on tracked agent tubes using Alg. \ref{alg:ped-move}.} Our experiments show that the model is not only capable of handling tube representations of an object's track but also extracting crucial information from them for classifying action.  

To better analyze the effects of DROI Alignment as opposed to other improvements, our method specifically improves the ability of our model to classify action classes with higher dynamic change. In Figure \ref{fig:classification_ablation}, the addition of DROI Alignment provides significant improvements to classifying actions like Turns, Crosses, and Movement Direction. We observe that these actions naturally require more information on trajectory, so it makes sense that the inclusion of tube-based alignment provides our network with an improved ability to detect them.  

\subsection{Deployment of RALACs to Research Vehicle}
\label{section:deployment_to_bolty}

To enhance the scene understanding capabilities of the WATonomous research platform \cite{dempster2022drg}---and provide preliminary insight into the effects action classification can have in autonomous vehicle decision making---RALACs was deployed to the WATonomous perception stack, and integrated with the environment model. 

\subsubsection{Test Scenario}
Real-world testing was carried out at the Waterloo Region Emergency Services Training and Research Centre (WRESTRC). In this work, we present the effects of action classification on an edge-case scenario previously tasked to the WATonomous team by the organizers of the SAE Autodrive Challenge. As shown in Figure \ref{fig:action:env_model_deploy}'s lane graphs, the pedestrian is close enough to the road such that our perception system is unsure as to whether or not the pedestrian is on the road or not. Our vehicle's task is to proceed forward after determining that the pedestrian will not interfere with our path despite being so close to the road.

\begin{algorithm}[!t]
    
    \caption{Detection - Frame Buffer Time Sync}
    \begin{algorithmic}[1]
    \renewcommand{\algorithmicrequire}{\textbf{Input:}}
    \renewcommand{\algorithmicensure}{\textbf{Output:}}
    \Require Frame \textit{timestamps}, observed \textit{bboxes} \Comment{both sorted from most recent to oldest}
    \Ensure For each frame $timestamp$, finds the temporally closest $bbox$
    \State $match\_i \textcolor{changes}{\gets} 0$ \Comment{\textcolor{changes}{Index of bounding box to check match}}
    \State $matched\_bbs \textcolor{changes}{\gets} []$
    \For{$s \in timestamps$}
        \While{TRUE}
            \State $match\_\delta \textcolor{changes}{\gets} abs(s - bboxes[match\_i].s)$
            \State $next\_match\_\delta \textcolor{changes}{\gets} abs(s - bboxes[match\_i + 1].s)$
            \State $match\_i \textcolor{changes}{\gets} match\_i + 1$
            \If{$match\_\delta < next\_match\_\delta$} \Comment{\textcolor{changes}{Next time delta is greater than current time delta}}
                \State BREAK
            \EndIf

        \EndWhile
        \State $matched\_bbs.append(bboxes[match\_i-1])$
    \EndFor
    \State \Return $matched\_bbs$
    \end{algorithmic} 
    \label{alg:action:time_sync}
\end{algorithm}

\subsubsection{Perception Integration} 

In order to deploy RALACs to the WATonomous autonomous driving stack (ADS), the networks inputs need to be generated in a real-time manner, which is achieved by the data pipeline in Figure \ref{fig:action:deployment_pipeline}. 

In brief, the history of 2D detections is added to each track. The detections are then time synchronized (see Alg. \ref{alg:action:time_sync}) with a buffered temporal view from the camera stream. Both the temporal view and the tube of observation are then run through RALACs. The result is a binary classification for each class, and the class with the highest score (if it is over 0.5) is added to the action classification history for that track.


\begin{figure*} 
    \centering
  \subfloat[First frame in the view given to ACAR]{%
      \includegraphics[width=0.33\linewidth] {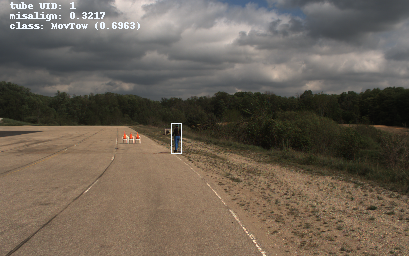}}
    \hfill
  \subfloat[Middle (keyframe) in the view given to ACAR]{%
        \includegraphics[width=0.33\linewidth]{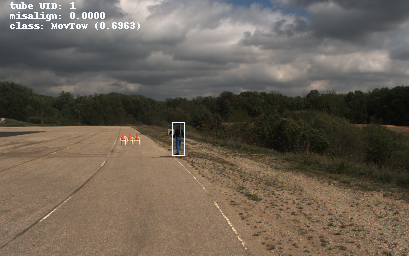}}
    \hfill
  \subfloat[Last frame in the view given to ACAR]{%
        \includegraphics[width=0.33\linewidth]{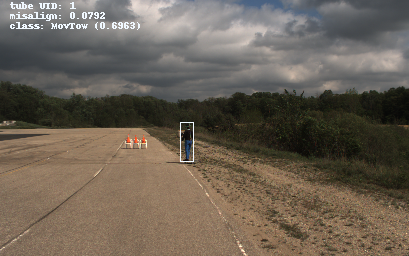}}
  \caption{Figure showing the visual and RoI inputs to the ACAR model, as well as the top classification score across the entire tube, and the temporal misalignment for the frame.}
  \label{fig:action:deployment_qualitative} 
\end{figure*}

\begin{figure*}
    \centering
  \subfloat[An overly conservative behavioral decision to stop given no visual indication that the pedestrian will interfere. Occurs when geometric heuristics specified in Alg. 2 of \cite{dempster2022drg} are used.]{%
       \includegraphics[width=0.49\linewidth,height=5cm]{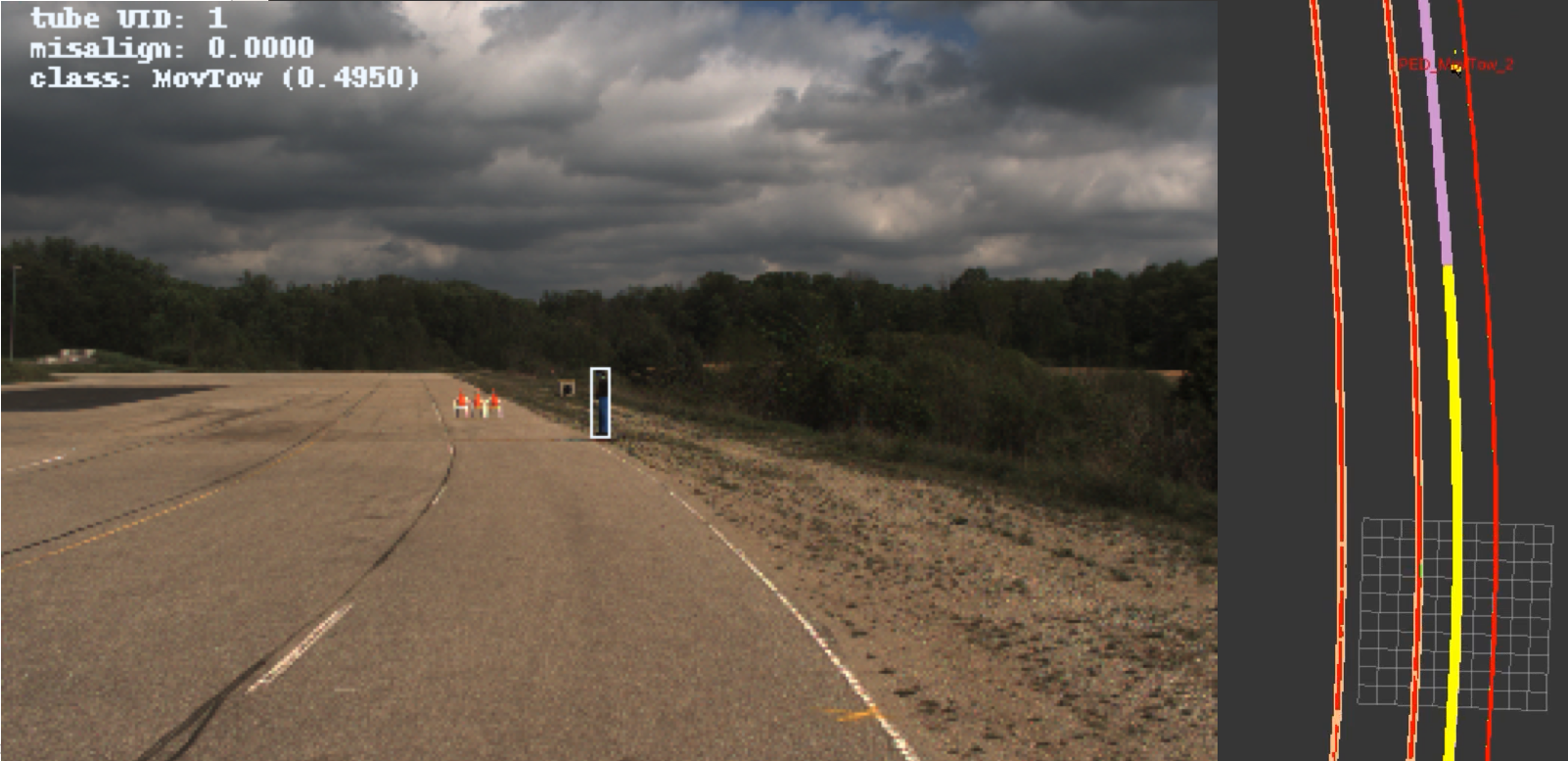}}
    \hfill
  \subfloat[Reasonable behavioral decision to drive past a stopped pedestrian, enabled by accurate action classification \textcolor{changes}{as specified in} Alg. \ref{alg:action:env_model_handle_ped}.]{%
        \includegraphics[width=0.49\linewidth,height=5cm]{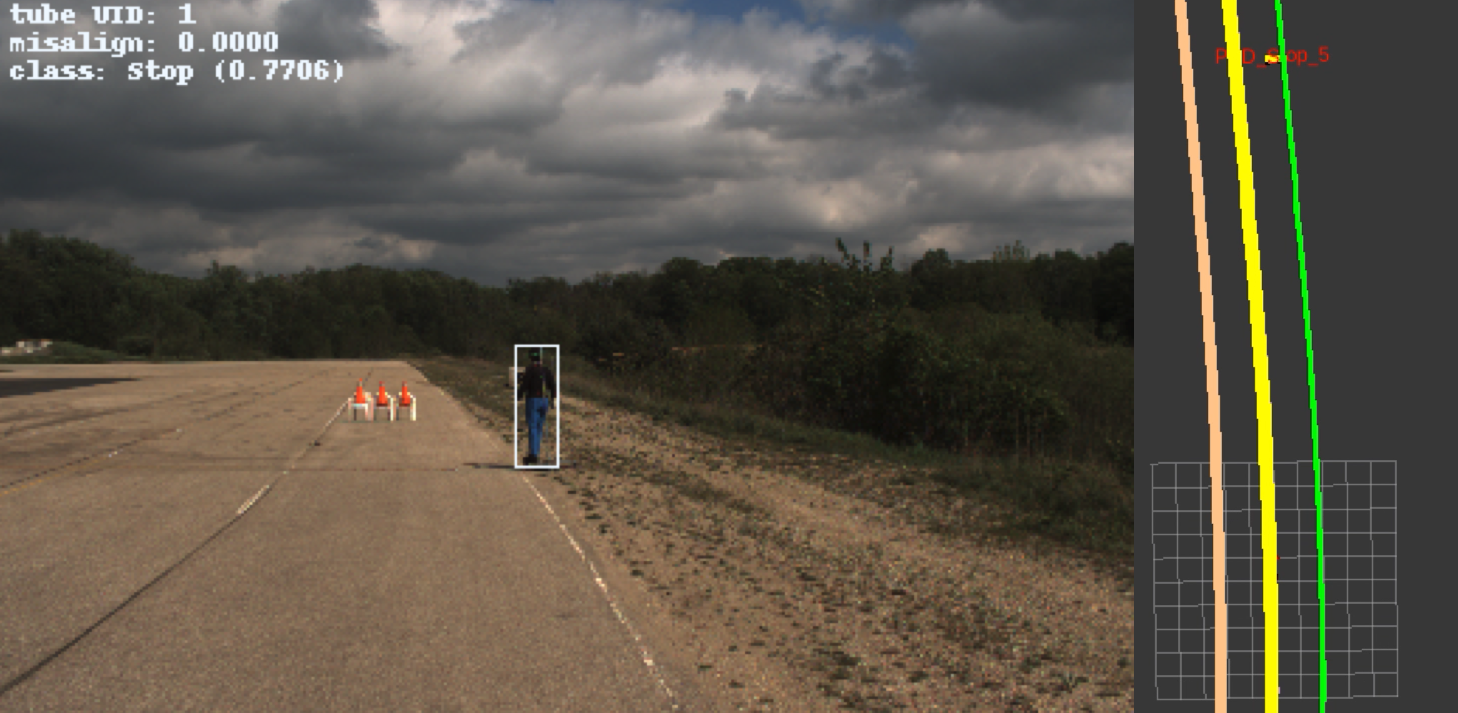}}
  \caption{Comparing different ego behavior in the stopped pedestrian scenario.}
  \label{fig:action:env_model_deploy} 
\end{figure*}

Figure \ref{fig:action:deployment_qualitative} presents the qualitative results of the deployment, showing the input and output of the deployed action classification model. As seen in the figure, the spatial footprint is tracked across the temporal view to provide a better visual description of the region of interest, as discussed in Section {\ref{section:action_class_w_tube_priors}. 

Table \ref{tab:action:deploy_quant_results} presents the quantitative results of the deployment, comparing the keyframe based model against the tube based model. As seen in the table, DROI Alignment produces a correct classification of \textit{stopped} the majority of the time, whereas a simple key-frame alignment scheme misclassifies the majority of the time. This can be attributed to the poor region of interest input to the action classification model, as analyzed before in Section {\ref{section:ablation}.

\begin{table}[h]
\begin{center}
\normalsize
\begin{tabular}{ l | c | c | c}
\hline
Model & Stop & MovTow & MovAway \\
\hline
Keyframe  &  \textbf{0.15} & 0.00 & 0.85  \\
DROI &  \textbf{0.54} & 0.26 & 0.20 \\
\hline
\end{tabular}
\vspace{1mm}
\caption{Ratio of classifications over the test scenario. As shown in the table, DROI-Alignment classifies the agent correctly for a longer period of time. Without the information of the pedestrian's track, the use of Keyframe alignment produces more false classifications. \textcolor{changes}{This movement} close to the lane graph would trigger our ego vehicle to stop out of uncertainty.}
\label{tab:action:deploy_quant_results}
\end{center}
\end{table}

\subsubsection{Integration with Environment Model}

To integrate action classification to the ASD environment model \cite{dempster2022drg}, incoming tracks of objects were simply appended with a history of action classifications. The environment model then uses this additional information to better discern whether or not to create relations between the actors and other entities in the relation graph. 

For example, refer to the previous modeling of pedestrian agents and their potential conflict relations with lanelets as laid out in Alg. 2 of \cite{dempster2022drg}. In that original naive implementation of the algorithm, the decision of whether or not a pedestrian conflicts with the lanelet was based on (1) whether the linear trajectory of the pedestrian intersects with the lanelet and (2) if the pedestrian is stationary, whether it is within some radius of the lanelet. It is clear that these geometric heuristic rules lead to a large number of false positive conflict relationships being instantiated. 


\begin{algorithm}[!t]
    
    \caption{Handle Pedestrian With Action}
    \begin{algorithmic}[1]
    \renewcommand{\algorithmicrequire}{\textbf{Input:}}
    \renewcommand{\algorithmicensure}{\textbf{Output:}}
    \Require Tracked \textit{ped} with history of actions, history time $t$ to consider, set of interference actions $int\_set$.
    \Ensure Boolean indicating if the pedestrian should be inserted into the conflict graph.
    \State $n\_interfere \textcolor{changes}{\gets} 0$ \Comment{\textcolor{changes}{Counter of interference actions}}
    \State $n\_total \textcolor{changes}{\gets} 0$ \Comment{\textcolor{changes}{Counter of total actions considered}} 
    \For{$action \in ped.action\_hist$}
        \State $now \textcolor{changes}{\gets} time.now()$ \Comment{\textcolor{changes}{Get current time}}
        \If{$now - action.stamp > t$} 
             \State BREAK 
        \EndIf
        \State \textcolor{changes}{$n\_total \leftarrow n\_total + 1$}
        \If{$action \in int\_set$} \Comment{\textcolor{changes}{If action is interference}}
            \State $n\_interfere \textcolor{changes}{\gets} n\_interfere + 1$ 
        \EndIf
    \EndFor
    \State \Return $n\_interfere / n\_total > 0.5$ \Comment{\textcolor{changes}{Interfere threshold}}
    \end{algorithmic} 
    \label{alg:action:env_model_handle_ped}
\end{algorithm}

The action classification history helps to limit these false positive cases by classifying the intentions of pedestrians based on their and the scene's appearance descriptions, rather than their geometrical descriptions on the lane graph. In Figure \ref{fig:action:env_model_deploy} we see an example of this, comparing the resulting relationship graph under the geometrical implementation, versus the graph under the action classification implementation. The difference in the conflict creation algorithm is presented in Alg. \ref{alg:action:env_model_handle_ped}, \textcolor{changes}{which determines whether a pedestrian's action history interferes with the route. This is done via a simple threshold of the ratio of actions which fall within an interference set $int\_set$ to total actions within a time history $t$.} In the test scenario $int\_set = \{Stop, Wait2X\}$ and $t=5s$ \textcolor{changes}{were} used.

Overall, this scenario provides important insight into the usefulness of action classification when lane graph heuristics fail. Without information that the pedestrian is perfectly stationary, our vehicle can never make a confident decision to proceed forward. In future works, we hope to extend our experiments to encompass more scenarios and edge cases.

\section{Conclusion}
\label{section:conclusion}

In this work, we develop methods for online action classification in road settings that focus on the identification of active agents and the use of agent-action contexts to determine their actions. We demonstrate that our contributions and proposed architecture can surpass the baseline performance of 3D-RetinaNet on the ROAD dataset and we tackle the deployment challenges of using such a system on a real vehicle platform. 

Towards the future, we welcome more datasets that set the standard for action recognition by including more varied and better defined agent actions. In addition, we believe that future efforts should explore the uses of action classification in decision making, and provide further analysis of its benefits. We believe that it may also be possible to achieve additive performance benefits from existing approaches by combining our contributions with existing post-processing ideas that work well, such as the de-duplication approach from the Argus++ authors.

\section*{Acknowledgment}

This work was supported by NSERC CRD 537104-18, in partnership with General Motors Canada. We would also like to thank Chuan Tian (Ben) Zhang for his early ideas and the WATonomous server infrastructure team whose support made this work possible.
%
\bibliographystyle{unsrt}
\bibliography{manuscript}

\end{document}